\documentclass{bmvc2k}

\title{SlotDiT: Object-Centric Representations for Diffusion Transformers}

\addauthor{Gjergj Plepi}{plepi@ais.uni-bonn.de}{1}
\addauthor{Sven Behnke}{behnke@cs.uni-bonn.de}{1}

\addinstitution{
 Autonomous Intelligent Systems, \\
 Computer Science Institute VI -- \\ Intelligent Systems and Robotics, \\
 Center for Robotics and the Lamarr \\ Institute for Machine Learning and \\ Artificial Intelligence, \\ University of Bonn, Germany
}

\runninghead{Plepi \& Behnke}{SlotDiT: Object-Centric Diffusion Transformer}

\usepackage{amssymb}
\usepackage{amsmath}
\DeclareMathOperator*{\softmax}{softmax}

\newcommand{\norm}[1]{\left\|#1\right\|}     
\newcommand{\E}{\mathbb{E}}                  

\newcommand{\NoiseLevel}{k}                 	
\newcommand{\NumDiffSteps}{K}               	
\newcommand{\AlphaBar}[1]{\bar{\alpha}_{#1}}	
\newcommand{\Noise}{\boldsymbol{\epsilon}}  	
\newcommand{\NoiseT}[1]{\boldsymbol{\epsilon}_{#1}} 
\newcommand{\Velocity}{\mathbf{v}}          	
\newcommand{\VelocityT}[1]{\mathbf{v}_{#1}}     
\newcommand{\Model}{\mathbf{v}_{\theta}}        
\newcommand{\NoiseLevels}{\mathbf{k}}           
\newcommand{\NoisySlotsT}[2]{\mathbf{S}_{#1}^{#2}}    
\newcommand{\NoisySlotsMany}[3]{\mathbf{S}_{#1:#2}^{#3}} 
\newcommand{\LossWeight}[1]{w(#1)}              

\newcommand{\Method}{\text{SlotDiT}}
\newcommand{\sdvae}{\text{SD-VAE}}
\newcommand{\imagevae}{\text{ImageVAE}}
\newcommand{\sdvaeDiT}{\text{DiT + SD-VAE}}
\newcommand{\imagevaeDiT}{\text{DiT + ImageVAE}}
\newcommand{\vavae}{\text{VA-VAE}}
\newcommand{\vavaeDiT}{\text{DiT + VA-VAE}}
\newcommand{\videovae}{\text{VideoVAE}}
\newcommand{\videovaeDiT}{\text{DiT + VideoVAE}}
\newcommand{\raeDiT}{\text{DiT + DINOv2}}
\newcommand{\invdyn}{\text{IDM}}
\newcommand{\nonoc}{Non-OC}

\newcommand{\ocwm}{\text{OC-WM}}
\newcommand{\gtslots}{\text{GT slots}}

\newcommand{\blockfour}{\textsc{Block-4}}
\newcommand{\blockeight}{\textsc{Block-8}}

\newcommand{\taskbb}{\textit{block-to-block}}
\newcommand{\taskbrl}{\textit{block-to-relative-location}}
\newcommand{\taskbbrl}{\textit{block-to-block-relative-location}}

\newcommand{\cliport}{\text{CLIPort}}
\newcommand{\ltsyn}{\text{LanguageTable-Synthetic}}
\newcommand{\ltreal}{\text{LanguageTable-Real}}
\newcommand{\bridge}{\text{Bridgev2}}

\newcommand{\NumPreds}{T} 
\newcommand{\NumFrames}{\tau} 
\newcommand{\NumBuffer}{N} 
\newcommand{\NumHistoryFrames}{M} 
 
\newcommand{\Caption}{\mathcal{C}}
\newcommand{\TextEmbs}{\mathbf{C}}

\newcommand{\ImageRange}[2]{\mathbf{X}_{#1:{#2}}}

\newcommand{\PredImageRange}[2]{\hat{\mathbf{X}}_{#1:{#2}}}
\newcommand{\ImageT}[1]{\textbf{X}_{#1}}
\newcommand{\PredImageT}[1]{\hat{\textbf{X}}_{#1}}

\newcommand{\DinoFeatsT}[1]{\textbf{h}_{#1}}
\newcommand{\PredDinoFeatsT}[1]{\hat{\textbf{h}}_{#1}}

\newcommand{\NumSlots}{N_\Slots}
\newcommand{\SlotDim}{D}
\newcommand{\Slots}{\mathbf{S}}
\newcommand{\SlotsT}[1]{\textbf{S}_{#1}}

\newcommand{\PredSlotsT}[1]{\hat{\textbf{S}}_{#1}}
\newcommand{\SingleSlot}{\textbf{s}}
\newcommand{\SingleSlotsT}[2]{\textbf{s}_{#1}^{#2}}

\newcommand{\SlotsMany}[2]{\mathbf{S}_{#1:#2}}
\newcommand{\PredSlotsMany}[2]{\hat{\mathbf{S}}_{#1:#2}}

\newcommand{\FeatureMapsT}[1]{\textbf{h}_{#1}}
\newcommand{\DimFeats}{D_h}
\newcommand{\NumLocs}{L}

\newcommand{\GRU}{\text{GRU}}
\newcommand{\Attention}{\textbf{A}}
\newcommand{\OutSlotAttention}{\textbf{U}}

\newcommand{\NumPredLayers}{N_\text{P}}
\newcommand{\TokenDim}{D_\text{Pred}}

\def\R{{\mathbb R}}

\newcommand{\Loss}{\mathcal{L}}

\newcommand{\dn}{$\downarrow$}                  
\newcommand{\up}{$\uparrow$}                    
\newcommand{\horizon}[2]{$#1\!\to\!#2$}         
\newcommand{\best}[1]{\textbf{#1}}              
\newcommand{\secondbest}[1]{\underline{#1}}     
\newcommand{\TBD}{---}                          

\usepackage{booktabs}
\usepackage{multirow}
\usepackage{colortbl}
\usepackage{url}

\usepackage{tikz}
\usetikzlibrary{arrows.meta}  %
\usepackage{adjustbox}

\begin{document}

\maketitle

\begin{abstract}
Text-conditioned latent diffusion models perform strongly in video generation
and are promising backbones for robotic applications.
However, existing approaches rely on pixel-level or VAE-based latent representations that lack explicit semantic structure, leaving the impact of the representation space largely unexplored.
Slot-based object-centric representations offer a structured alternative by decomposing scenes into object-level latents, or \emph{slots}.
While they have shown success in dynamics modeling and planning, they have not yet been explored for diffusion-based generative modeling.
We introduce \Method{}, a text-guided Diffusion Transformer (DiT) that operates in a slot-based latent space.
Given a reference image and a language instruction, \Method{} decomposes the scene into  object-centric slots representing individual entities.
Conditioned on the instruction and observed scene context, 
the model autoregressively denoises future slot trajectories to predict scene dynamics.
To systematically investigate latent-space design for diffusion transformers, we compare slot-based representations against VAE-based and semantics-aligned alternatives within a unified DiT framework.
Our experiments show that using slots as DiT latents yields competitive video generation quality while consistently improving task-completion rates %
across four robotic datasets.
Furthermore, their compact representation provides a computationally efficient alternative to VAE-based and semantics-aligned latent spaces.
Overall, our results demonstrate that object-centric structure is a powerful inductive bias for diffusion-based generative modeling in robotic environments.
The project page is available at \url{https://slot-dit.github.io/}.

\end{abstract}

\section{Introduction}
\vspace{-0pt}
\label{sec:intro}

\begin{figure*}[t]
  \centering
  \begin{tikzpicture}[every node/.style={inner sep=0pt, outer sep=0pt}]
    \node[anchor=south west] (pipeline) at (0, 0)
      {\includegraphics[width=0.95\linewidth]{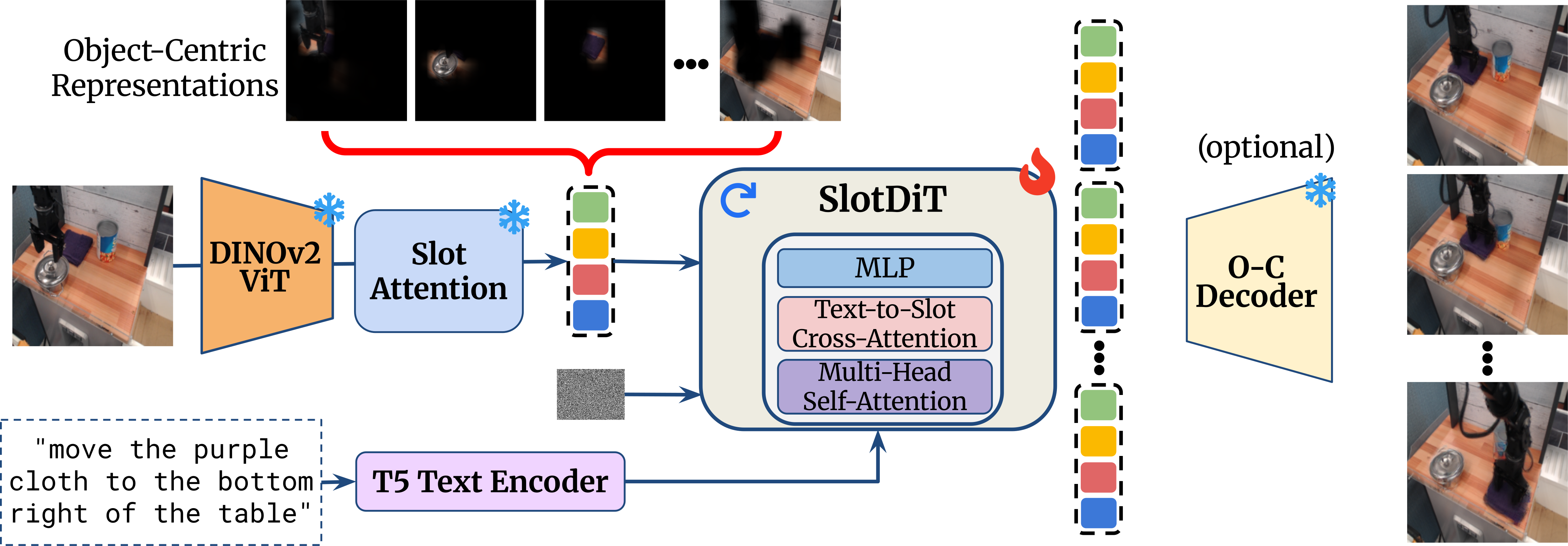}};

    \begin{scope}[
      x={(pipeline.south east)},
      y={(pipeline.north west)}
    ]
      \node[anchor=north] at (0.07, 0.34)
        {\small $\ImageT{1}$};
      \node[anchor=south] at (0.22, 0.)
        {\small $\Caption$};
      \node[anchor=south] at (0.58, 0.12)
        {\small $\TextEmbs$};
      \node[anchor=south] at (0.41, 0.39)
        {\small $\SlotsT{1}$};
       \node[anchor=south] at (0.285, 0.255)
        {\small noise vectors};
      \node[anchor=south] at (0.738, 0.78)
      {\small $\PredSlotsT{2}$};
      \node[anchor=south] at (0.738, 0.48)
      {\small $\PredSlotsT{3}$};
      \node[anchor=south] at (0.755, 0.1)
      {\small $\PredSlotsT{\NumPreds+1}$};
      \node[anchor=south] at (0.88, 0.78)
      {\small $\PredImageT{2}$};
      \node[anchor=south] at (0.88, 0.48)
      {\small $\PredImageT{3}$};
      \node[anchor=south] at (0.87, 0.1)
      {\small $\PredImageT{\NumPreds+1}$};
    \end{scope}

    \node[anchor=south west, font=\small\bfseries]
      at ([yshift=-8pt]pipeline.north west) {(a)};

    \draw[gray!55, densely dotted, line width=0.9pt]
      ([yshift=-0.2cm]pipeline.south west) --
      ([yshift=-0.2cm]pipeline.south east);

    \node[anchor=north east] (graph)
      at ([yshift=-0.7cm]pipeline.south east)
      {\includegraphics[width=0.38\linewidth]{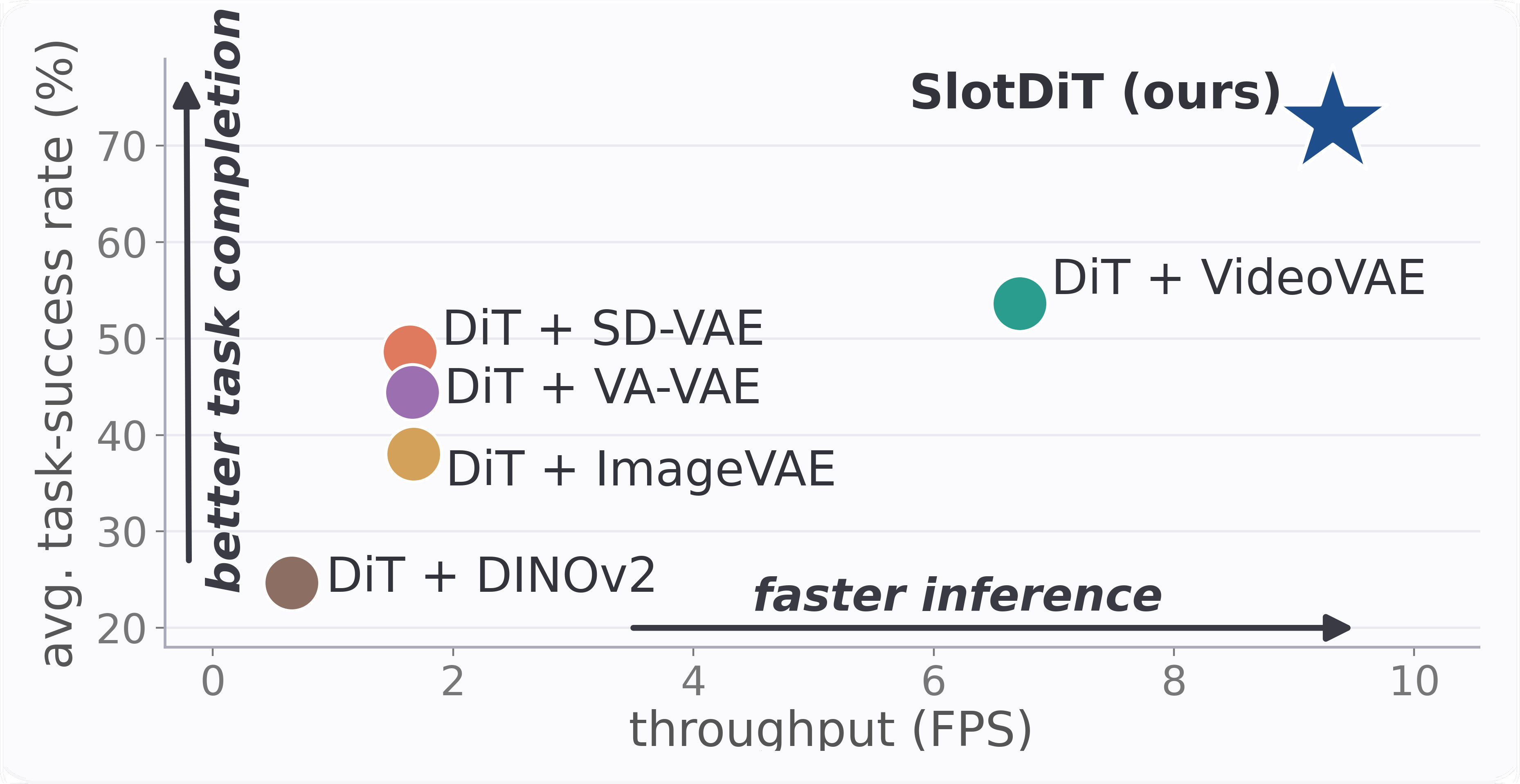}};
       \node[anchor=south, font=\small]
      at ([yshift=1.5pt]graph.north)
      {Efficiency vs. Task Success};

    \node[anchor=south west] (rep)
      at ([yshift=-2.97cm] pipeline.south west)
      {\includegraphics[width=0.56\linewidth]{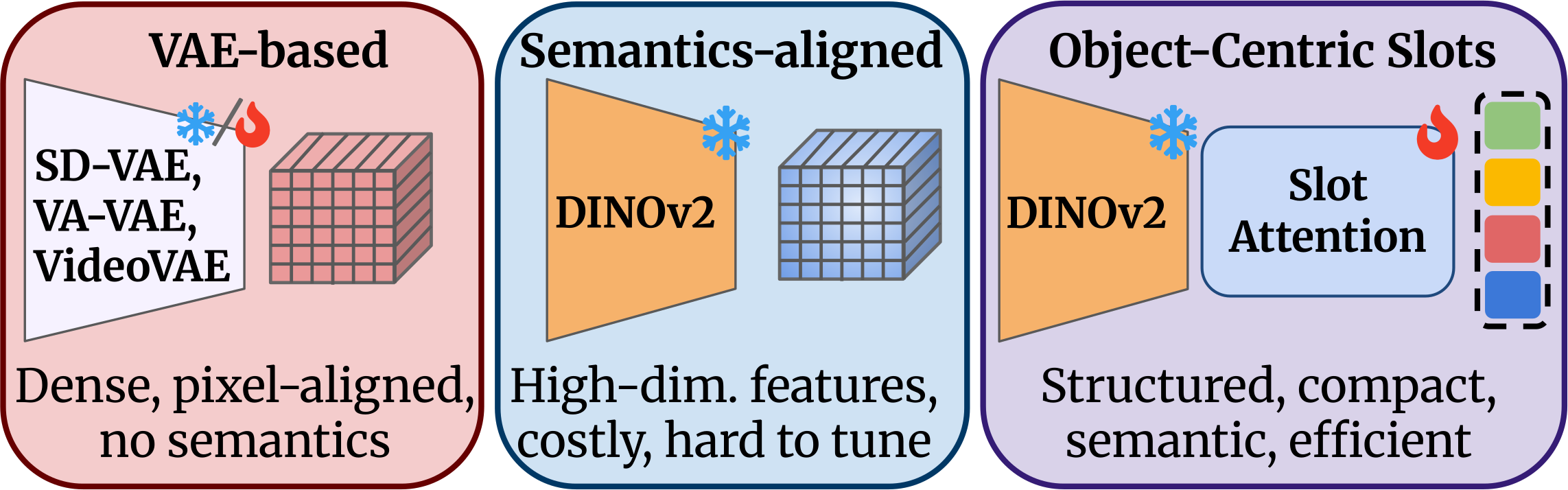}};
    \node[anchor=south, font=\small]
      at ([yshift=3pt]rep.north)
      {Same DiT, different latent representations};

    \node[anchor=south west, font=\small\bfseries]
      at ([yshift=3pt]rep.north west) {(b)};
    \node[anchor=south west, font=\small\bfseries]
      at ([xshift=2pt, yshift=1.8pt]graph.north west) {(c)};
  \end{tikzpicture}
  \caption{%
    Overview of \Method{}.
    \textbf{(a)}
    Given a reference image $\ImageT{1}$ and text instruction $\Caption$, \Method{} parses the scene into an object-centric slot representation $\SlotsT{1}$.
    Conditioned on the encoded instruction and context slots, an object-centric diffusion transformer (DiT) autoregressively denoises future slot trajectories
    $\PredSlotsT{2}, \ldots, \PredSlotsT{\NumPreds+1}$, 
    optionally decoded into future video frames $\PredImageT{2},\, \ldots,\, \PredImageT{\NumPreds+1}$.
    \textbf{(b)}
    Comparison of \Method{}'s slot-based latent
    space against established representation spaces used in DiTs.
    \textbf{(c)}
    Across four robotic datasets, \Method{} achieves higher task-completion rates while remaining substantially more efficient than the baselines.
  }
  \label{fig:teaser}
  \vspace{-3pt}
\end{figure*}

Diffusion models have shown remarkable success in image and video generation and have since been adapted to various other settings.
In robotics, their generative capabilities and probabilistic formulation have enabled applications in planning \cite{du2024video, ajay2023compositional}, policy learning
\cite{du2023learning, ko2024learning, jang2025dreamgen}, and world
modeling \cite{zhou2024robodreamer, ye2026world}.
State-of-the-art diffusion models typically use a transformer backbone \cite{Vaswani_AttentionIsAllYouNeed_2017, peebles2023scalable}, conditioning
signals such as text \cite{kong2024hunyuanvideo, yang2025cogvideox}, and, crucially, a learned latent space \cite{rombach2022high}.
While improved architectures and model and data scaling have driven substantial
progress, the latent representation space has received comparatively little
attention.
This gap is especially critical in robotic settings, where the latent
space directly determines what information is preserved for downstream
prediction, planning, and control \cite{jha2026reconstruction}.

The latent spaces used by diffusion models are mostly learned with VAE-based encoders, with Stable Diffusion VAE (\sdvae{}) \cite{rombach2022high} being the most prominent example.
Subsequent works extend this paradigm in various directions,
including semantics-aligned VAEs~\cite{yao2025reconstruction},
joint optimization of the VAE and the latent diffusion model~\cite{leng2025repa}, 
and spatio-temporal video compression~\cite{gupta2024photorealistic, wan2025wan}.
Despite these advances, such approaches remain fundamentally reconstruction-driven, prioritizing high visual fidelity over semantic structure.
Thus, it remains unclear whether these latent spaces are well suited for robotic applications, where successful task completion matters more than pixel-level reconstruction.

To address this limitation, recent works \cite{zheng2026diffusion} replace the VAE with large pretrained visual foundation models \cite{oquab2023dinov2, tschannen2025siglip}, resulting in a more semantically aligned latent space.
However, training latent diffusion transformers in these high-dimensional feature spaces is not trivial and requires additional architectural adaptations and training heuristics to be effective \cite{zhang2025both}.

In parallel, object-centric representations \cite{Locatello_ObjectCentricLearningWithSlotAttention_2020}---which decompose scenes into compact sets of object-level latent vectors---have proven beneficial across various robotic tasks, including dynamics modeling \cite{villartextocvp} and continuous control \cite{villar2025playslot, daniel2026latent, spieler2026slot}, suggesting that object-centric structure may be a useful inductive bias for generative modeling in robotic environments.
However, their use as latent representations for diffusion remains unexplored.

In this work, we propose \Method{}, illustrated in Figure~\ref{fig:teaser}a, a text-guided Diffusion Transformer that operates within an object-centric slot-based latent space.
Given a single reference image and a language instruction, \Method{} first parses the scene into a set of object-centric slots.
Conditioned on these slots and the encoded instruction, the model then iteratively denoises Gaussian noise to autoregressively generate future slot trajectories, which can optionally be decoded into future video frames.

Unlike dense VAE-based latents or high-dimensional semantics-aligned representations, the slot-based latent space is compact and structured, representing each frame with only a small number of slot tokens (Figure~\ref{fig:teaser}b).
To assess the impact of this design, we conduct a controlled comparison of latent spaces for diffusion transformers, evaluating slot-based, VAE-based, and semantics-aligned representations within a single, unified DiT framework \cite{song2025historyguided} where all models share the same architecture, training procedure, and inference scheme, and differ only in the underlying latent space.

We evaluate \Method{} on text-guided video generation across four robotic
datasets spanning simulated and real-world environments, and on robot control
in two simulated environments. %
Our experiments show that slot-based representations achieve competitive video generation quality while consistently yielding higher task-success rates than the alternative latent spaces.
Notably, strong visual-quality metrics do not necessarily translate into successful task completion: VAE-based baselines often have better perceptual scores, yet perform substantially worse than \Method{} on instruction following and task-solving.
Furthermore, \Method{} outperforms all baselines in %
the robot control evaluation, and its compact object-centric latent space yields
substantial efficiency gains over other representations (Fig.~\ref{fig:teaser}c).

In summary, our main contributions are:
\vspace{-5pt}
\begin{itemize}
	\setlength{\itemsep}{2pt}
	\setlength{\parskip}{0pt}
	\setlength{\parsep}{0pt}
	\item We introduce \Method{}, a text-guided Diffusion Transformer that operates in an object-centric latent space, enabling object-centric generative modeling in robotic settings.
	\item We present a controlled study of latent spaces for diffusion transformers, comparing slot-based, VAE-based and semantics-aligned latents within a unified framework.
	\item We show that slot-based representations consistently improve downstream task-success rates in text-guided video generation and robot control, while providing a substantially more efficient latent space than VAE-based and semantics-aligned alternatives.
\end{itemize}

\section{Related Work}
\label{sec:related_work}

\paragraph{Object-Centric Representation Learning:}
Object-centric representation methods aim to decompose a scene into individual object components. 
There are different approaches to representing these objects, including patch-based representations \cite{lin2020space}, particle-based representations \cite{daniel2022unsupervised}, explicit object prototypes \cite{monnier2020deep, villar2021unsupervised}, and slot-based latents \cite{Locatello_ObjectCentricLearningWithSlotAttention_2020}, where each individual scene entity is encoded into a distinct latent vector, called slot.

The \emph{Slot Attention} module \cite{Locatello_ObjectCentricLearningWithSlotAttention_2020}
is the core building block of most slot-based representation learning methods, mapping image features into a set of object slots.
Subsequent works extended this framework to video inputs \cite{Kipf_ConditionalObjectCentricLearningFromVideo_2022, elsayed2022savi++, Singh_STEVE_2022}.
Recently, significant progress has been made to scale slot-based representation learning to complex real-world scenes through key advancements such as leveraging pretrained visual feature extractors \cite{Seitzer_BridgingTheGapToRealWorldObjectCentricLearning_2023, aydemir2023self}, improving feature-level learning \cite{kakogeorgiou2024spot, Zadaianchuk_VideoSaur_2024, manasyan2025temporally},
or employing diffusion decoders \cite{slotdiffusion, wu2023slotdiffusion, yemez2025slot, nguyen2026improved}.

Many works \cite{villar2023object, Wu_SlotFormer_2022, villartextocvp} have leveraged slots for video prediction, modeling spatio-temporal object dynamics via autoregressive transformer predictors trained using image and slot forecasting objectives.
Additionally, slot-based dynamics models have been proven beneficial for multiple downstream tasks such as policy learning~\cite{Mosbach_SOLDReinforcementLearningSlotObjectCentricLatentDynamics, villar2025playslot, daniel2026latent}, Model Predictive Control (MPC)~\cite{spieler2026slot, nam2026causal} or visual reasoning~\cite{nam2026causal}.
In contrast to prior work, we explore slot-based representations as latent spaces for diffusion transformers, enabling object-centric diffusion-based generative modeling in robotic settings.

\vspace{-0pt}
\paragraph{Representation Space of Diffusion Models:}
Diffusion models \cite{sohl2015deep, ho2020denoising} have emerged as a leading paradigm for image and video generation \cite{rombach2022high, esser2024scaling}.
However, operating them directly in pixel space is computationally expensive, motivating latent diffusion models (LDMs) \cite{rombach2022high, blattmann2023align}, which perform the diffusion process in a compressed learned latent space.
These latent spaces are traditionally learned with reconstruction-based objectives using pretrained Variational Autoencoders (VAEs) \cite{kingma2013auto}, which prioritize visual fidelity over semantic structure.
Following the Stable Diffusion VAE (\sdvae) \cite{rombach2022high, esser2024scaling}, many works propose improved latent representations. \vavae{} \cite{yao2025reconstruction} aligns \sdvae{} latents with vision foundation models \cite{oquab2023dinov2}, REPA~\cite{leng2025repa} enables joint optimization with LDMs, and \videovae{} encodes videos spatio-temporally \cite{yu2024language, gupta2024photorealistic, yang2025cogvideox, wan2025wan}.
Despite these advances, such latent spaces remain largely reconstruction-oriented and lack explicit semantic structure.

Recent works \cite{zheng2026diffusion, tong2026scaling} explore replacing VAEs with Representation Autoencoders (RAE), leveraging pretrained representation encoders \cite{oquab2023dinov2, tschannen2025siglip} to obtain a semantically aligned latent space.
While these approaches improve the learned representations, the resulting high-dimensional features require additional architectural modifications and training heuristics to make them suitable for diffusion modeling \cite{zhang2025both}.
In contrast to existing works, we explore structured object-centric latent spaces for diffusion transformers, highlighting their effectiveness and computational efficiency in robotic environments.

\vspace{-0pt}
\paragraph{Text-guided Video Generation:}
Textual instructions provide external guidance for video generation, providing information about objects and their intended motions.
Many works have utilized Transformer-based predictors for text-guided video generation \cite{hu2022make, villartextocvp, fu2023tell}.
Specifically, recent approaches \cite{villartextocvp, jeong2025object} use slot-based representations and show the benefits of a structured latent space for text-guided manipulation in robotic environments. 

Recently, video diffusion models \cite{ho2022video} have demonstrated strong capabilities for generating high-quality videos conditioned on a text prompt \cite{kong2024hunyuanvideo, yang2025cogvideox, gupta2024photorealistic, agarwal2025cosmos, chen2024gentron, ma2025latte}.
Their success mostly stems from leveraging pretrained image diffusion models \cite{singer2023makeavideo}, scaling training data \cite{blattmann2023stable}, or having large Diffusion Transformer backbones \cite{peebles2023scalable}. 
Due to their capabilities, text-guided diffusion models have also been applied in various downstream tasks in robotic scenarios.
However, current video diffusion models still operate mostly in a reconstruction-aligned latent space, leveraging either a pretrained \sdvae{} \cite{chen2024gentron, ma2025latte} or a \videovae{} \cite{yang2025cogvideox, gupta2024photorealistic, ali2025world}, focusing mainly on high-fidelity generation.

\section{Methodology}

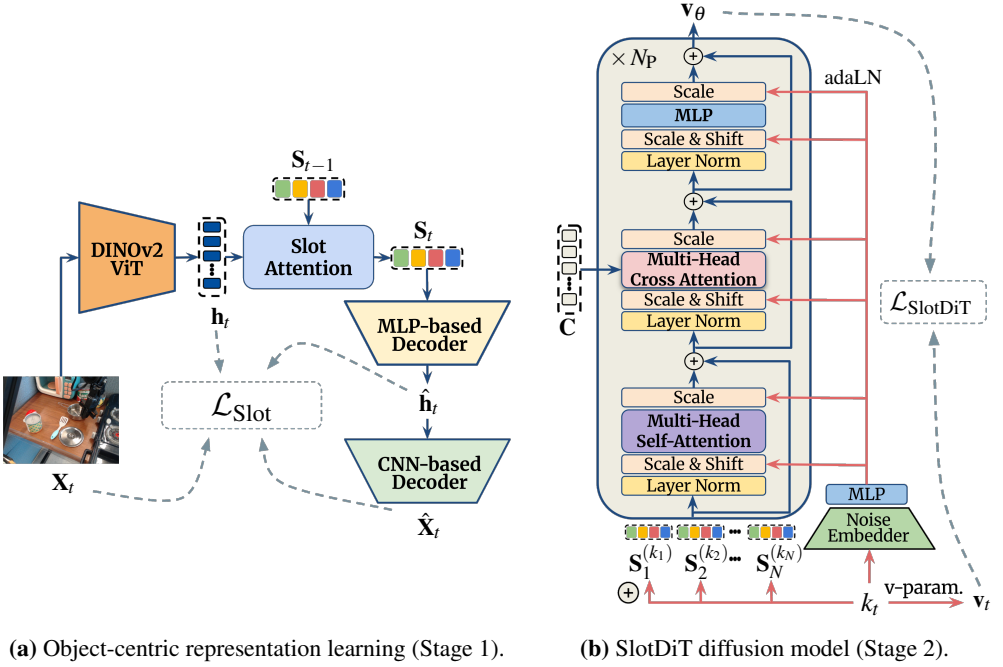
\begin{figure*}[t]
	\centering
	\begin{minipage}[b]{0.52\linewidth}
		\centering
		\begin{minipage}[b][0.5783\textwidth][c]{\linewidth}
		\centering
		\begin{tikzpicture}
			\node(oc)[anchor=south west, inner sep=0pt, outer sep=0pt] at (0, 0)
				{\includegraphics[width=\linewidth]{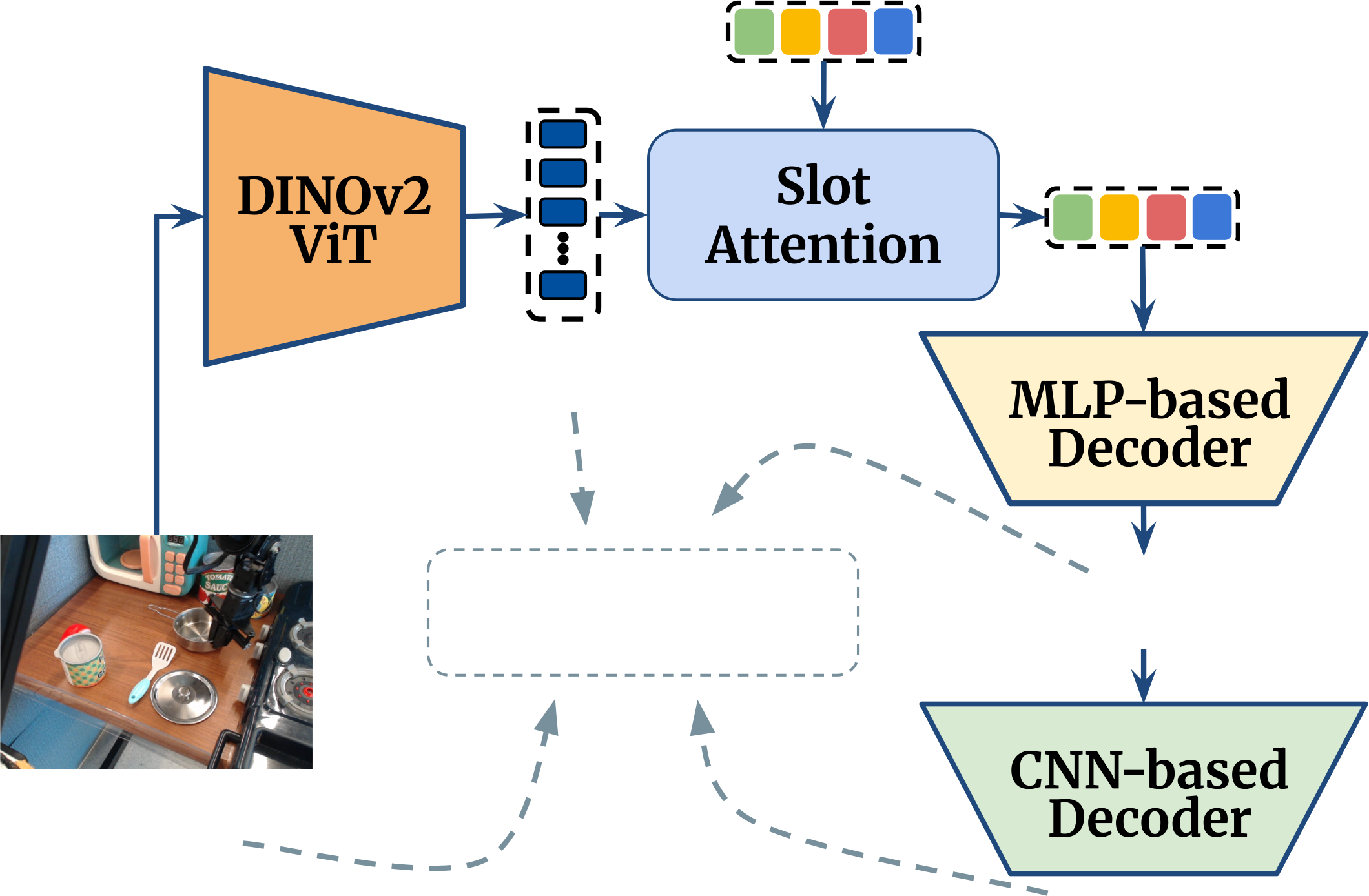}};
			\begin{scope}[x={(oc.south east)}, y={(oc.north west)}]
				\node[anchor=north] at (0.12, 0.14)
					{\small $\ImageT{t}$};
				\node[anchor=south] at (0.43, 0.52)
					{\small $\DinoFeatsT{t}$};
				\node[anchor=west] at (0.55, 1.05)
					{\small $\SlotsT{t-1}$};
				\node[anchor=south] at (0.83, 0.78)
					{\small $\SlotsT{t}$};
				\node[anchor=south] at (0.84, 0.26)
					{\small $\PredDinoFeatsT{t}$};
				\node[anchor=north] at (0.84, 0.02)
					{\small $\PredImageT{t}$};
				\node at (0.47, 0.31)
					{\large $\Loss_{\text{Slot}}$};
			\end{scope}
		\end{tikzpicture}
		\end{minipage}
		\\[50pt]
		{\small \textbf{(a)}~Object-centric representation learning (Stage~1).}
	\end{minipage}
	\hfill
	\begin{minipage}[b]{0.44\linewidth}
		\centering
		\begin{tikzpicture}
			\node(dit)[anchor=south west, inner sep=0pt, outer sep=0pt] at (0, 0)
				{\includegraphics[width=\linewidth]{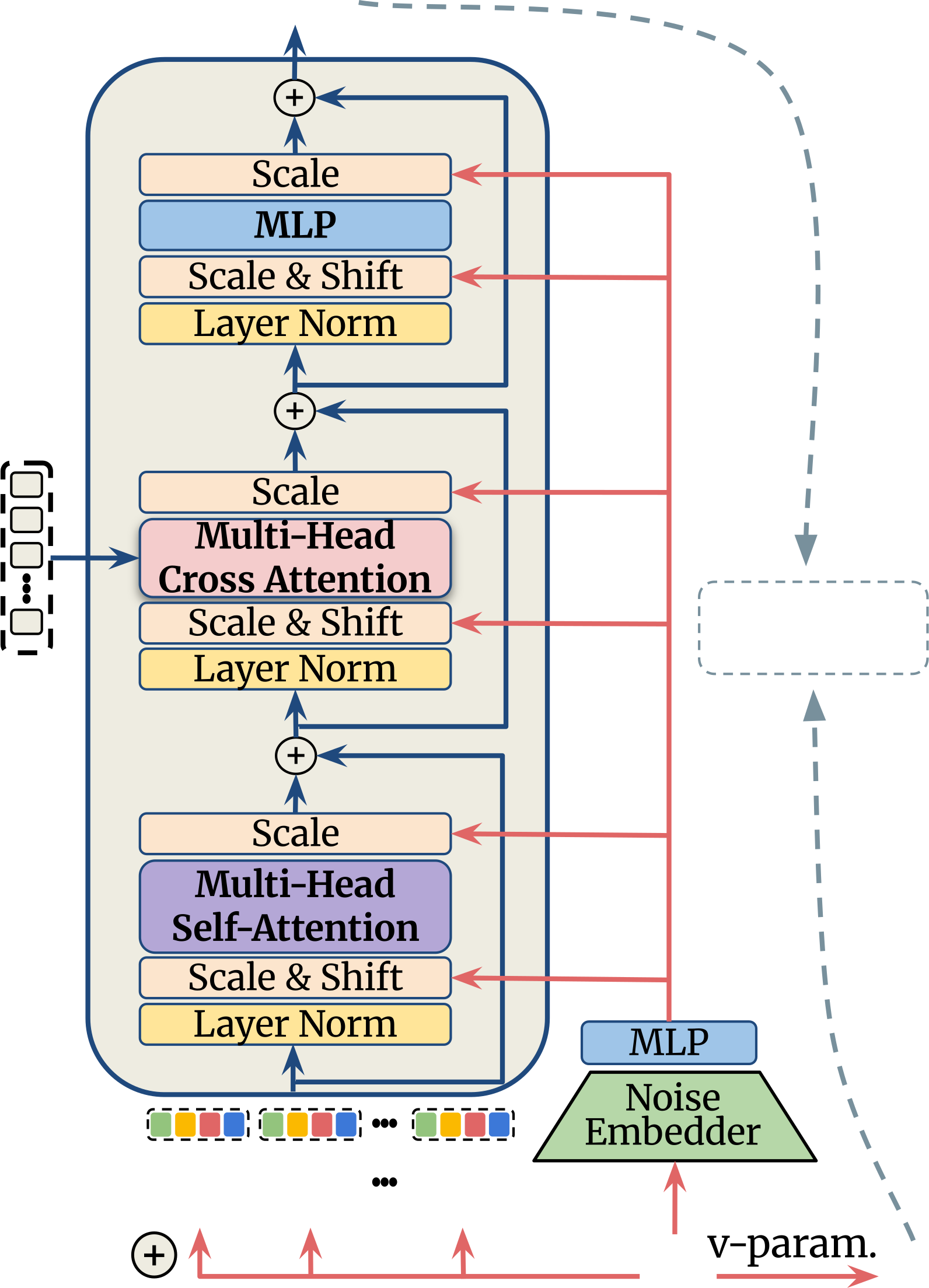}};
			\begin{scope}[x={(dit.south east)}, y={(dit.north west)}]
				\node[anchor=west] at (0.265, 0.999)
					{$\Model$};
				\node[anchor=west] at (0.1, 0.92)
					{\small $\times\,\NumPredLayers$};
				\node[anchor=south west] at (-0.018, 0.44)
					{\small $\TextEmbs$};
				\node[anchor=west] at (0.6, 0.89)
					{\footnotesize adaLN};
				\node[anchor=north] at (0.22, 0.122)
					{\small $\NoisySlotsT{1}{(\NoiseLevel_1)}$};
				\node[anchor=north] at (0.35, 0.122)
					{\small $\NoisySlotsT{2}{(\NoiseLevel_2)}$};
				\node[anchor=north] at (0.52, 0.122)
					{\small $\NoisySlotsT{\NumBuffer}{(\NoiseLevel_{\NumBuffer})}$};
				\node[anchor=north] at (0.73, 0.04)
					{$\NoiseLevel_t$};
				\node[anchor=north] at (0.99, 0.035)
				{\small $\textbf{v}_t$};
				\node[anchor=south] at (0.87, 0.475)
					{$\Loss_{\Method}$};
			\end{scope}
		\end{tikzpicture}
		{\small \textbf{(b)}~SlotDiT diffusion model (Stage~2).}
	\end{minipage}
	\caption{
		\Method{} training.
		\textbf{(a)}~\emph{Slot representation learning:}
		A frozen DINOv2-ViT extracts visual features $\DinoFeatsT{t}$ from each	frame $\ImageT{t}$, which are decomposed into object-centric slots $\SlotsT{t}$ via Slot Attention.
		The model is trained by minimizing a feature and image reconstruction objective $\Loss_{\text{Slot}}$.
		\textbf{(b)}~\emph{SlotDiT:}
		With the slot encoder frozen, noisy slot sets $\NoisySlotsT{t}{(\NoiseLevel_t)}$ and text embeddings $\TextEmbs$ are	processed by a Diffusion Transformer.
		Per-frame noise levels are injected via adaLN, and the model is trained
		with the diffusion objective $\Loss_{\Method}$.
	}
	\label{fig:slotdit_arch}
	\vspace{-0.cm}
\end{figure*}

We propose \Method{}, depicted in Figure~\ref{fig:slotdit_arch}, a latent Diffusion Transformer (DiT) that learns in a slot-based object-centric latent space.
Given a single reference image $\ImageT{1}$ and a language instruction $\Caption$, \Method{} first encodes the scene into a set $\SlotsT{1}$ of $\NumSlots{}$ object-centric latent vectors, called slots (Sec.~\ref{sub: object_centric_module}).
Then, conditioned on the instruction and observed scene context, the model iteratively denoises future slot trajectories from Gaussian noise to generate $\NumBuffer-1$ future sets of slots.
The last $\NumHistoryFrames$ generated slot sets are recursively used as context for the next prediction step, enabling autoregressive generation of the subsequent $\NumPreds$ sets of slots $\PredSlotsMany{2}{\NumPreds+1}$ (Sec.~\ref{sub: diffusion_transformer}). 
Optionally, the generated slots can be decoded into future image frames $\PredImageRange{2}{\NumPreds+1}$.

\vspace{0pt}
\subsection{Learning Slot-Based Representations} %
\label{sub: object_centric_module}

Our object-centric representation learning module is based on the DINOSAUR framework \cite{Seitzer_BridgingTheGapToRealWorldObjectCentricLearning_2023}.
Following \cite{villartextocvp}, we extend it to the video domain by incorporating a Transformer~\cite{Vaswani_AttentionIsAllYouNeed_2017} encoder as a temporal transition function and an image decoder to enable pixel-level decoding.
Given an input video sequence $\ImageRange{1}{\NumFrames}$, this module parses each frame into a set of object-centric slots $\SlotsT{t} = (\SingleSlotsT{t}{1}, \ldots, \SingleSlotsT{t}{\NumSlots})$, for $t \in \{1, \ldots, \NumFrames\}$, where every individual slot $\SingleSlot \in \R^{\SlotDim}$ represents a single scene entity.

For each frame $\ImageT{t}$, a DINOv2 vision transformer (DINOv2-ViT) \cite{oquab2023dinov2} encodes the frame into a feature map $\FeatureMapsT{t} \in \R^{\NumLocs \times \DimFeats}$.
A Slot Attention module \cite{Locatello_ObjectCentricLearningWithSlotAttention_2020} then iteratively refines the previous slots $\SlotsT{t-1}$ by attending to the visual features, allowing different slots to specialize on different scene entities.
Slots attend to the visual features via cross-attention, with the attention weights normalized over the slot axis to enforce competition among slots for representing each feature location:
\vspace{-0pt}
\begin{align}
	\Attention = \softmax_{\NumSlots} \left( \frac{q(\SlotsT{t-1})\, k(\FeatureMapsT{t})^{\top}}{\sqrt{\SlotDim}} \right) \in \R^{\NumSlots \times \NumLocs},
	\label{eq:slot_attn}
\end{align}
where $q$ and $k$ are learned linear projections. 
The attention weights are then normalized across feature locations to compute a weighted mean of the projected features, which is then used to update the slots through a Gated Recurrent Unit (GRU) \cite{Cho_GRU_2014}:
\vspace{-0pt}
\begin{align}
	W_{i,j} = \frac{\Attention_{i,j}}{\sum_{l=1}^{\NumLocs}\Attention_{i,l}}, \qquad
	\OutSlotAttention = W\, v(\FeatureMapsT{t}), \qquad
	\SlotsT{t} = \GRU(\OutSlotAttention, \SlotsT{t-1}),
	\label{eq:slot_update}
\end{align}
where $v$ is a learned linear projection, and $\SlotsT{t}$ are the resulting slots for frame $\ImageT{t}$.

To compose the scene from the parsed object-centric representations, we employ a two-stage decoder similar to that of~\cite{villartextocvp}.
First, an MLP-based broadcast decoder independently maps each slot to an object feature map and an alpha mask.
After mask normalization across slots, feature maps $\PredDinoFeatsT{t} \in \R^{\NumLocs \times \DimFeats}$ are reconstructed via a weighted sum.
Finally, a CNN-based image decoder maps the reconstructed features to render the video frame $\PredImageT{t}$.

\subsection{Slot-based Text-Conditioned Diffusion Transformer}
\label{sub: diffusion_transformer}
Given the initial set of slots $\SlotsT{1}$ and the language instruction $\Caption$, \Method{} autoregressively generates future slot trajectories through an iterative denoising process in the latent space~\cite{rombach2022high}. 
The language instruction is first encoded into a sequence of text token embeddings $\TextEmbs$ using a pretrained T5 encoder \cite{raffel2020exploring}.
These embeddings provide the semantic and motion information that conditions the generation process.

\Method{} operates on a fixed-length temporal window of $\NumBuffer$ slot sets.
At each generation step $t$, \Method{} receives as input the most recent $\NumHistoryFrames$ context slots $\SlotsMany{t-\NumHistoryFrames+1}{t}$, the encoded text instruction $\TextEmbs$, and random Gaussian noise for the remaining $\NumBuffer-\NumHistoryFrames$ future frames in the window.
The context slots and text embeddings are projected to the model token
dimensionality $\TokenDim$ and jointly processed by the Diffusion
Transformer backbone.
Conditioned on both the observed slot history and language
instruction, \Method{} iteratively denoises the noisy future slots to
predict the next $\NumBuffer-\NumHistoryFrames$ slot sets.
The last $\NumHistoryFrames$ generated slot sets are then reused as
context for the subsequent prediction step, allowing the model to
autoregressively generate the full future slot sequence.

The latent denoising process is parameterized by a Diffusion Transformer
architecture operating over the slot representations and conditioned on text tokens.
We design \Method{} as a Diffusion Transformer \cite{peebles2023scalable} composed of $\NumPredLayers$ identical blocks, illustrated in Figure~\ref{fig:slotdit_arch}b.
Each block consists of a bidirectional multi-head self-attention layer,
a text-to-slot cross-attention layer, and an MLP, where every component
is preceded by LayerNorm and wrapped in a residual connection.
Following prior works~\cite{perez2018film,	peebles2023scalable}, we use adaptive layer-normalization (adaLN) to inject the per-frame noise-level embeddings into every block.

The self-attention layer jointly processes all slots within the temporal window, modeling spatio-temporal relations between object-centric representations across frames.
We apply rotary embeddings (RoPE)~\cite{su2024roformer} as positional encoding and study two distinct variants.
The first preserves slot permutation-equivariance by applying RoPE only along the temporal axis, sharing the same rotation across all slots within a frame, whereas the second applies RoPE along both temporal and slot axes, breaking permutation-equivariance.
The text-to-slot cross-attention layer then incorporates semantic and motion-related
information from the text embeddings into the slots.
Finally, the MLP processes each token independently.

\vspace{-0pt}
\subsection{Model Training and Inference}
\label{sub: training_inference}
We train \Method{} following a two-stage training procedure, depicted in Figure~\ref{fig:slotdit_arch}.
First, we train the object-centric representation module.
We then train \Method{} in the resulting slot latent space while keeping
the slot encoder frozen.

\vspace{-0pt} 
\paragraph{Learning Slots:} 
The object-centric decomposition and decoding modules are trained using a combined image- and feature-reconstruction objective over the input sequence:
\vspace{-0pt}
\begin{align}
	\Loss_{\text{Slot}} = \frac{1}{\NumFrames} \sum_{t=1}^{\NumFrames} \left( \norm{\PredImageT{t} - \ImageT{t}}_2^2 + \norm{\PredDinoFeatsT{t} - \DinoFeatsT{t}}_2^2 \right),
	\label{eq:slot_loss}
\end{align}
\vspace{-0pt}
where $\NumFrames$ is the length of the training sequence, $\PredDinoFeatsT{t}$ and $\PredImageT{t}$ are the reconstructed features and frame, and $\DinoFeatsT{t}$ and $\ImageT{t}$ are their targets.

\vspace{-0pt}
\paragraph{Training \Method{}:}
We train \Method{} following the Diffusion Forcing Transformer (DFoT) framework~\cite{song2025historyguided}.
Each training example consists of $\NumBuffer$ slot sets $\SlotsMany{1}{\NumBuffer}$ parsed from a video, together with its language instruction $\Caption$ encoded into text embeddings $\TextEmbs$.
To enable classifier-free guidance~\cite{ho2022classifier}, the text embeddings $\TextEmbs$ are randomly replaced with a learned null embedding during training.  %

Following the noise-as-masking paradigm~\cite{chen2024diffusion}, each frame is corrupted with an \emph{independent} diffusion noise level $\NoiseLevel \in \{0, \ldots, \NumDiffSteps-1\}$, where $\NumDiffSteps$ denotes the number of diffusion steps and $\AlphaBar{\NoiseLevel}$ the cumulative coefficient of the noise schedule.
This formulation allows every token to contribute to the training objective, thus improving token utilization, while also enabling a variable number of history frames during inference.

Formally, given a per-frame noise level $\NoiseLevel_t$ and Gaussian noise $\NoiseT{t} \sim \mathcal{N}(\mathbf{0}, \mathbf{I})$, the diffusion forward process corrupts the set of slots of each frame independently:
\vspace{-0pt}
\begin{align}
	\NoisySlotsT{t}{(\NoiseLevel_t)} = \sqrt{\AlphaBar{\NoiseLevel_t}}\, \SlotsT{t} + \sqrt{1 - \AlphaBar{\NoiseLevel_t}}\, \NoiseT{t} .
	\label{eq:forward_diffusion}
\end{align}
\vspace{-0pt}
The noisy slot window $\NoisySlotsMany{1}{\NumBuffer}{(\NoiseLevels)}$, together with the per-frame noise levels $\NoiseLevels = (\NoiseLevel_1, \ldots, \NoiseLevel_{\NumBuffer})$ and text embeddings $\TextEmbs$, is passed to \Method{}, which learns the reverse denoising process.
We adopt the $\Velocity$-prediction parameterization \cite{salimans2022progressive}, where the target for frame $t$ is
\vspace{-0pt}
\begin{align}
	\VelocityT{t} = \sqrt{\AlphaBar{\NoiseLevel_t}}\, \NoiseT{t} - \sqrt{1 - \AlphaBar{\NoiseLevel_t}}\, \SlotsT{t} .
	\label{eq:v_target}
\end{align}
\vspace{-0pt}
We denote by $\Model$ the velocity predicted by \Method{}, and train the model to minimize the reweighted denoising objective
\vspace{-0pt}
\begin{align}
	\Loss_{\Method} = \E_{\SlotsMany{1}{\NumBuffer},\, \TextEmbs,\, \NoiseLevels,\, \Noise} \left[ \frac{1}{\NumBuffer} \sum_{t=1}^{\NumBuffer} \LossWeight{\NoiseLevel_t}\, \norm{ \Model\!\left( \NoisySlotsMany{1}{\NumBuffer}{(\NoiseLevels)}, \NoiseLevels, \TextEmbs \right)_{t} - \VelocityT{t} }_2^2 \right] ,
	\label{eq:dfot_loss}
\end{align}
\vspace{-0pt}
where $\Model(\cdot)_{t}$ denotes the predicted velocity for frame $t$ and $\LossWeight{\NoiseLevel_t}$ is a per-frame loss weight.
Following~\cite{hang2023efficient, chen2024diffusion}, we
use fused min-SNR loss reweighting.  %

\vspace{-0pt}
\paragraph{Inference:}
At inference, \Method{} generates slots autoregressively in temporal
windows, conditioning on one observed slot set initially and the most
recent $\NumHistoryFrames$ predictions thereafter, initializing the remaining positions with Gaussian noise.
The full window is then iteratively denoised using a DDIM sampler \cite{song2021denoising}.
At each step, \Method{} predicts $\mathbf{v}_{\theta}$, %
from which the next, less noisy slots are computed, while the slot history remains unchanged. 

To improve sample quality and instruction following, we guide the generation with classifier-free guidance \cite{ho2022classifier}, extending the history guidance formulation \cite{song2025historyguided} to the text condition.
Specifically, we compute both a conditional prediction---using the clean history slots and text instruction---as well as an unconditional prediction, where the history slots are replaced with noise and the text embeddings with a learned null embedding.
The two predictions are combined using a single guidance scale that encourages consistency with both the observed history and the language instruction.
After denoising, the last $\NumHistoryFrames$ generated slot sets are reused as history context for the next temporal window.
This autoregressive process is repeated until all $\NumPreds$ future slot sets are generated.

\vspace{-0pt}
\section{Experiments}
\vspace{-0pt}
To evaluate \Method{}, we investigate three key research questions: 
(i) can diffusion transformers operate effectively in an object-centric
latent space?
(ii) do slot-based representations improve task-solving and
motion-following capabilities in robotic environments compared to
established latent spaces?
and (iii) do slots offer a more computationally efficient representation?
To this end, we evaluate \Method{} on two downstream tasks:
text-guided video generation and robot control.
The former assesses the visual fidelity of the generated futures, while the latter evaluates the instruction following capability and usefulness of the learned representations for planning and task execution.
We conduct a controlled comparison across four robotic datasets,
isolating the effect of the underlying representation space.

\vspace{-0pt}
\subsection{Experimental Setup}
 
\vspace{-0pt}
\paragraph{Datasets:}
We evaluate \Method{} on four distinct language-conditioned robotic datasets:
the synthetic tabletop manipulation environments
\ltsyn{} \cite{lynch2023interactive} and \cliport{} \cite{shridhar2022cliport},
the real-robot tabletop dataset \ltreal{}~\cite{lynch2023interactive}, and BridgeData V2
(\bridge{}) \cite{walke2023bridgedata}, which features diverse household
manipulation tasks performed by a service robot.
Together, they span both simulated and real-world robotic
environments of varying complexity and contain tasks specified through
natural-language instructions.
We evaluate text-guided video generation and task execution on all four datasets, and robot
control on LanguageTable-Synthetic and \cliport{}.
Additional dataset details are provided in Appendix~\ref{appendix:datasets}.

\vspace{-0pt}
\paragraph{Evaluation Metrics:}
To assess video generation performance, we report LPIPS~\cite{lpips},
FVD \cite{fvd}, and JEDi~\cite{jedi}.
LPIPS measures the perceptual similarity between generated and ground-truth frames, whereas FVD and JEDi are video-level metrics that compare the distributions of real and generated videos, reflecting temporal consistency and motion realism.
JEDi tends to remain more reliable on the relatively small evaluation sets we use.

As these metrics do not capture instruction following or task completion, we additionally report task-success rate, where success is determined directly by the environment for the robot control task, or computed using a VLM-as-judge for text-guided video generation. 
Specifically, a VLM jointly processes the text instruction and a subset of generated frames to predict whether the task was successfully completed.
The task-success rate is the percentage of sequences classified as successful.
We use Qwen3-VL-30B-A3B-Instruct~\cite{bai2025qwen3} as the judge.

\vspace{-0pt}
\paragraph{Baselines:}
We evaluate \Method{} against two groups of baselines.
The first group consists of slot-based models that operate on the same object-centric representations as \Method{}, but employ autoregressive Transformer predictors instead of diffusion-based generative modeling.
Namely, we compare against
TextOCVP~\cite{villartextocvp}, an object-centric Transformer model for
language-conditioned video prediction.
For robot control, we additionally compare against
\ocwm{}~\cite{jeong2025object}, a slot-based world model with a
Transformer dynamics predictor.

The second group isolates the effect of the representation space.
Here, all models use the same DiT architecture, similar training procedure, and
inference scheme as \Method{}, but operate in different latent spaces.
We consider VAE-based latent spaces, including
\sdvae{}~\cite{rombach2022high} (\sdvaeDiT{}),
a custom \imagevae{} (\imagevaeDiT{}),
a pretrained VA-VAE \cite{yao2025reconstruction} (\vavaeDiT{}),
and a \videovae{} \cite{lin2024open} (\videovaeDiT{}) that applies a spatio-temporal encoding to the input video.
Additionally, we consider a semantics-aligned RAE-style latent space built from DINOv2 features~\cite{oquab2023dinov2, zheng2026diffusion}.

\vspace{-3pt}
\paragraph{Implementation Details.}

We train all models on two NVIDIA A6000 (48GB) GPUs, using 
a frozen T5-small text encoder for language conditioning.
The object-centric decomposition module processes images at a resolution of
$224\times224$ ($336\times336$ for \cliport{}).
Scenes are represented using
$\NumSlots \in \{8,10\}$ slots of dimension
$\SlotDim \in \{128,256\}$.
The Diffusion Transformer uses
$\NumPredLayers=8$ blocks ($\NumPredLayers=14$ blocks on \bridge{}) with hidden dimension $\TokenDim=512$, and is trained on temporal windows of $\NumBuffer=10$ slot sets.
We train using the DFoT framework with a cosine noise schedule over $\NumDiffSteps = 1000$ diffusion steps, the $\Velocity$-prediction parameterization \cite{salimans2022progressive}, fused min-SNR loss reweighting, and DDIM sampling~\cite{song2021denoising} at inference.

To ensure a controlled comparison of representation spaces, all DiT
baselines share the same architecture, training procedure, and
inference setup as \Method{}, differing only in the underlying latent
space.
Complete architectural details, hyperparameters, and training settings
for all models are provided in Appendix~\ref{appendix:implementation_details}.

\vspace{-0pt}
\subsection{Results}
\label{sec:results}

\begin{table*}[t]
	\centering
	\footnotesize
	\setlength{\tabcolsep}{2.5pt}
	\renewcommand{\arraystretch}{1.0}
	\begin{tabular}{@{}l l ccc ccc ccc c@{}}
		\toprule
		Dataset & Model
		& \multicolumn{3}{c}{\horizon{1}{9}}
		& \multicolumn{3}{c}{\horizon{1}{19}}
		& \multicolumn{3}{c}{\horizon{1}{29}}
		& Success \\
		\cmidrule(lr){3-5}\cmidrule(lr){6-8}\cmidrule(lr){9-11}
		&   & LPIPS\dn & FVD\dn & JEDi\dn
		& LPIPS\dn & FVD\dn & JEDi\dn
		& LPIPS\dn & FVD\dn & JEDi\dn
		& \%\up \\
		\midrule
		\multirow{7}{*}{\cliport{}}
		& \sdvaeDiT{}      & 0.138 & 165.01 & 3.02 & 0.153 & 138.23 & 3.17 & 0.163 & 159.37 & 3.82 & 36.6\% \\
		& \imagevaeDiT{}   & 0.121 & 155.37 & 2.57 & 0.138 & 183.05 & 3.14 & 0.153 & 217.46 & 2.82 & 25.8\% \\
		& \vavaeDiT{}      & 0.153 & 193.88 & 3.43 & 0.181 & 241.57 & 4.60 & 0.189 & 211.48 & 6.14 & 39.0\% \\
		& \videovaeDiT{}   & \best{0.055} & \best{24.78} & \best{0.46} & 0.098 & \best{71.82} & \best{1.49} & \best{0.069} & \best{30.08} & \best{0.50} & 57.1\% \\
		& \raeDiT{}        & 0.184 & 252.57 & 4.18 & 0.213 & 298.37 & 6.25 & 0.221 & 226.38 & 8.10 & 12.2\% \\
		\rowcolor{gray!8}
		& TextOCVP         & \secondbest{0.062} & 96.28 & 1.36 & \secondbest{0.078} & 122.09 & 2.23 & 0.098 & 150.52 & 2.17 & \secondbest{63.0\%} \\
		\rowcolor{gray!8}
		& \Method{} (Ours) & 0.066 & \secondbest{83.66} & \secondbest{1.13} & \best{0.077} & \secondbest{79.27} & \secondbest{1.73} & \secondbest{0.092} & \secondbest{90.37} & \secondbest{1.85} & \best{87.8\%}  \\
		\midrule
		\multirow{7}{*}{LT-Syn{}}
		& \sdvaeDiT{}      & \best{0.088} & \secondbest{29.36} & 1.83 & 0.102 & \secondbest{36.54} & 2.06 & \secondbest{0.117} & \secondbest{38.98} & 1.83 & 49.2\% \\
		& \imagevaeDiT{}   & 0.095 & 35.27 & \secondbest{1.30} & 0.108 & 38.80 & \secondbest{1.09} & 0.126 & 42.40 & \secondbest{0.94} & 43.1\% \\
		& \vavaeDiT{}      & \secondbest{0.089} & 53.57 & 3.42 & \secondbest{0.100} & 67.97 & 2.95 & \secondbest{0.117} & 65.89 & 3.80 & 44.2\% \\
		& \videovaeDiT{}   & 0.090 & \best{26.36} & \best{1.15} & \best{0.087} & \best{30.57} & \best{0.71} & \best{0.111} & \best{32.37} & \best{0.67} & \secondbest{55.8\%} \\
		& \raeDiT{}        & 0.114 & 64.59 & 2.79 & 0.125 & 88.21 & 2.86 & 0.137 & 105.71 & 3.23 & 14.7\% \\
		\rowcolor{gray!8}
		& TextOCVP         & 0.104 & 66.17 & 2.29 & 0.112 & 92.20 & 3.43 & 0.152 & 150.34 & 4.01 & 25.9\% \\
		\rowcolor{gray!8}
		& \Method{} (Ours) & 0.110 & 70.49 & 1.36 & 0.119 & 81.76 & 1.55 & 0.131 & 83.34 & 1.48 & \best{79.7\%} \\
		\bottomrule
	\end{tabular}
	\caption{
		Text-guided video generation on the \cliport{} and \ltsyn{} synthetic datasets.
		We report visual fidelity metrics (LPIPS, FVD, and JEDi) at three prediction horizons, while the rightmost column measures whether the generated video satisfies the language-conditioned task.
		Best results are shown in \best{bold}, second-best \secondbest{underlined}.
	}
	\label{tab:videogen_synthetic}
	\vspace{-0pt}
\end{table*}
\begin{table*}[t]
	\centering
	\footnotesize
	\setlength{\tabcolsep}{2.5pt}
	\renewcommand{\arraystretch}{1.0}
	\begin{tabular}{@{}l l ccc ccc ccc c@{}}
		\toprule
		Dataset & Model
		& \multicolumn{3}{c}{\horizon{1}{9}}
		& \multicolumn{3}{c}{\horizon{1}{19}}
		& \multicolumn{3}{c}{\horizon{1}{29}}
		& Success \\
		\cmidrule(lr){3-5}\cmidrule(lr){6-8}\cmidrule(lr){9-11}
		&   & LPIPS\dn & FVD\dn & JEDi\dn
		& LPIPS\dn & FVD\dn & JEDi\dn
		& LPIPS\dn & FVD\dn & JEDi\dn
		& \%\up \\
		\midrule
		\multirow{7}{*}{LT-Real}
		& \sdvaeDiT{}      & 0.113 & 61.21 & \best{1.29} & 0.133 & 89.26 & \secondbest{1.53} & 0.145 & 86.36 & \secondbest{1.52} & 59.0\% \\
		& \imagevaeDiT{}   & \secondbest{0.102} & \secondbest{50.37} & 2.32 & \secondbest{0.117} & \secondbest{77.99} & 2.78 & 0.130 & \secondbest{83.95} & 3.22 & 48.9\% \\
		& \vavaeDiT{}      & 0.108 & 69.14 & 1.98 & 0.124 & 114.03 & 2.00 & 0.138 & 116.04 & 2.40 & 51.9\% \\
		& \videovaeDiT{}   & \best{0.096} & \best{24.82} & \secondbest{1.51} & \best{0.100} & \best{34.69} & \best{1.38} & \best{0.125} & \best{32.34} & \best{1.05} & \secondbest{61.1\%} \\
		& \raeDiT{}        & 0.165 & 131.24 & 6.08 & 0.177 & 191.18 & 6.50 & 0.186 & 215.80 & 7.60 & 47.1\% \\
		\rowcolor{gray!8}
		& TextOCVP         & 0.104 & 157.60 & 7.67 & 0.119 & 250.69 & 8.34 & \secondbest{0.128} & 285.25 & 10.09 & 47.4\% \\
		\rowcolor{gray!8}
		& \Method{} (Ours) & 0.119 & 72.99 & 4.45 & 0.134 & 95.35 & 3.42 & 0.143 & 98.41 & 3.81 & \best{64.6\%} \\
		\midrule
		\multirow{7}{*}{\bridge{}}
		& \sdvaeDiT{}      & 0.121 & \secondbest{51.38} & \secondbest{2.35} & 0.157 & \secondbest{74.58} & \secondbest{1.45} & 0.178 & \secondbest{94.91} & 1.87 & 49.7\% \\
		& \imagevaeDiT{}   & \best{0.096} & \best{48.53} & \best{1.72} & \best{0.136} & \best{65.97} & \best{1.21} & \best{0.158} & \best{87.75} & \best{1.49} & 34.4\% \\
		& \vavaeDiT{}      & 0.107 & 71.40 & 2.54 & 0.147 & 106.12 & 1.76 & 0.171 & 140.54 & \secondbest{1.77} & 42.7\% \\
		& \videovaeDiT{}   & \secondbest{0.102} & 57.25 & 4.36 & \secondbest{0.142} & 95.20 & 3.17 & \secondbest{0.162} & 108.32 & 2.80 & 40.6\% \\
		& \raeDiT{}        & 0.128 & 100.56 & 6.07 & 0.158 & 207.39 & 5.96 & 0.178 & 315.42 & 6.46 & 30.8\% \\
		\rowcolor{gray!8}
		& TextOCVP         & 0.146 & 164.83 & 2.77 & 0.168 & 210.77 & 2.57 & 0.189 & 263.06 & 2.30 & \best{63.4\%} \\
		\rowcolor{gray!8}
		& \Method{} (Ours) & 0.125 & 65.26 & 3.63 & 0.158 & 101.26 & 2.59 & 0.180 & 125.48 & 2.16 & \secondbest{57.6\%} \\
		\bottomrule
	\end{tabular}
	\caption{
		Text-guided video generation on the real-world datasets \ltreal{} and \bridge{}.
		We report visual fidelity metrics (LPIPS, FVD, and JEDi) at three prediction horizons, while the rightmost column measures whether the generated video satisfies the language-conditioned task.
		Best results are shown in \best{bold}, second-best \secondbest{underlined}.
	}
	\label{tab:videogen_real}
	\vspace{-0pt}
\end{table*}

\subsubsection{Text-Guided Video Generation.}
We report text-guided video generation results on \cliport{}, \ltsyn{}, Language\allowbreak Table-Real, and \bridge{} in Tables~\ref{tab:videogen_synthetic} and~\ref{tab:videogen_real}.
For each dataset, we show visual-quality metrics at prediction horizons $\NumPreds\in\{9,19,29\}$, 
and assess task completion using VLM-judged task-success rates on full-sequence predictions matching the ground-truth rollout length.

As shown in Table~\ref{tab:videogen_synthetic}, \Method{} achieves substantially better task-success rates than all baselines on both the \cliport{} and \ltsyn{} environments.
On \cliport{}, most VAE-based baselines perform poorly in both visual quality and task success, whereas \videovaeDiT{} achieves the strongest LPIPS, FVD, and JEDi scores, indicating high visual and temporal fidelity.
Nevertheless, \Method{} remains competitive in visual quality while
achieving substantially higher task-success rates.

We attribute this gap to the compact object-centric latent space, which represents each frame using only a few low-dimensional slot tokens.
While this design may limit the amount of visual detail that can be represented compared to high-capacity VAE latents, it provides a more efficient and structured representation for modeling task-relevant scene dynamics, as reflected by the high task-success rates of slot-based TextOCVP and \Method{} models.

A similar trend is observed on \ltsyn{}.
Although \imagevaeDiT{} and \videovaeDiT{} outperform \Method{} in visual-quality metrics, they achieve significantly lower task-success rates.
These results suggest that standard video-generation metrics do not fully capture task-solving and instruction-following performance in robotic environments: 
while visual fidelity is important, it does not necessarily correlate with successful task completion.
As shown in Figure~\ref{fig:videogen_qualitative_zoom}a, \videovaeDiT{} correctly identifies the target object but fails to perform the instruction, whereas \Method{} successfully
completes the task.

Table~\ref{tab:videogen_real} reports results for the real-world datasets. 
On \ltreal{}, \Method{} achieves the highest task-success rate, outperforming all alternative latent spaces and TextOCVP.
On \bridge{}, it remains among the strongest methods and outperforms the best VAE-based baseline by around 8 percentage points.
Consistent with the observations on the synthetic datasets, object-centric methods generally underperform in visual-quality metrics, yet consistently obtain higher task-success rates.
In particular, the strongest VAE-based models in visual quality, \videovaeDiT{} and \imagevaeDiT{}, achieve considerably lower task-success rates than \Method{}.
This further supports our observation that strong perceptual metrics do
not necessarily translate into successful task completion.
Figure~\ref{fig:videogen_qualitative_zoom}b shows a qualitative \bridge{} example in which \Method{} generates long-horizon rollouts that follow the language instruction and successfully execute the task, whereas baseline models fail.

\newlength{\zqw}\setlength{\zqw}{0.90cm}      %
\newlength{\zqh}\setlength{\zqh}{0.90cm}      %
\newlength{\zqsep}\setlength{\zqsep}{0.045cm} %
\newlength{\zqrowsep}\setlength{\zqrowsep}{0.05cm} %
\newlength{\zoomw}\setlength{\zoomw}{1.20cm}  %
\newlength{\zoomh}\setlength{\zoomh}{1.20cm}  %
\newlength{\zoomgap}\setlength{\zoomgap}{0.22cm} %
\newlength{\zoomvsep}\setlength{\zoomvsep}{0.08cm} %

\newcommand{\videoGenZoomLTSyn}[3]{%
  \begin{tikzpicture}[
    every node/.style={inner sep=0pt, outer sep=0pt},
    rowlabel/.style={anchor=east, font=\scriptsize, align=right},
    collabel/.style={anchor=south, font=\scriptsize},
    zlabel/.style={anchor=south, font=\scriptsize\itshape},
    instr/.style={anchor=south, font=\itshape\small, align=center,
                  text width=4.6cm},
    gtframe/.style={draw=black!35, line width=0.5pt, inner sep=0pt},
    zoomline/.style={line width=0.5pt, dashed},
  ]
    \foreach \c/\t in {0/1, 1/8, 2/15, 3/22, 4/29} {%
      \pgfmathsetlengthmacro{\xcol}{\c*(\zqw + \zqsep)}%
      \node[collabel] at (\xcol, 0.5\zqh + 0.04cm) {$t=\t$};
    }

    \pgfmathsetlengthmacro{\yrow}{0pt}%
    \pgfmathsetlengthmacro{\labelx}{-0.5*\zqw - 0.08cm}%
    \node[anchor=south west, font=\scriptsize] at (1.55*\labelx, -0.25\zqh + 0.16cm) {GT};
    \foreach \c/\fname in {0/1, 1/8, 2/15, 3/22, 4/29}{%
      \pgfmathsetlengthmacro{\xcol}{\c*(\zqw + \zqsep)}%
      \node at (\xcol, \yrow) {%
        \includegraphics[width=\zqw, height=\zqh]{#2/gt/#3/test_\fname.png}};
    }

    \pgfmathsetlengthmacro{\gc}{-1.5*(\zqh+\zqrowsep)}%
    \pgfmathsetlengthmacro{\xz}{4*(\zqw+\zqsep) + 0.5*\zqw + \zoomgap + 0.5*\zoomw}%
    \foreach \r/\modellabel/\modeldir/\modelcol in {%
        1/{\Method{}}/{slotdit}/{green!55!black},
        2/{DiT +\\ SD-VAE}/{sdvae}/{orange!85!black},
        3/{DiT +\\ VideoVAE}/{videovae}/{violet!80!black}%
    }{%
      \pgfmathsetlengthmacro{\yrow}{-\r*(\zqh + \zqrowsep)}%
      \pgfmathsetlengthmacro{\labelx}{(\zqw+\zqsep) - 0.5*\zqw - 0.08cm}%
      \node[rowlabel] at (\labelx, \yrow) {\modellabel};
      \foreach \c/\fname in {1/7, 2/14, 3/21}{%
        \pgfmathsetlengthmacro{\xcol}{\c*(\zqw + \zqsep)}%
        \node at (\xcol, \yrow) {%
          \includegraphics[width=\zqw, height=\zqh]{#2/\modeldir/#3/test_\fname.png}};
      }
      \pgfmathsetlengthmacro{\xlast}{4*(\zqw + \zqsep)}%
      \node[draw=\modelcol, line width=0.9pt, inner sep=0pt] (zsrc\r) at (\xlast, \yrow) {%
        \includegraphics[width=\zqw, height=\zqh]{#2/\modeldir/#3/test_28.png}};
      \pgfmathtruncatemacro{\zi}{\r-1}%
      \pgfmathsetlengthmacro{\yz}{\gc - (\zi-1)*(\zoomh+\zoomvsep)}%
      \node[draw=\modelcol, line width=0.9pt, inner sep=0pt] (zdst\r) at (\xz, \yz) {%
        \includegraphics[width=\zoomw, height=\zoomh]{#2/\modeldir/#3/test_28.png}};
      \draw[zoomline, \modelcol] (zsrc\r.north east) -- (zdst\r.north west);
      \draw[zoomline, \modelcol] (zsrc\r.south east) -- (zdst\r.south west);
    }

    \pgfmathsetlengthmacro{\xmid}{2.*(\zqw + \zqsep)}%
    \node[instr] at (\xmid, 0.5\zqh + 0.32cm) {``#1''};
  \end{tikzpicture}%
}

\newcommand{\videoGenZoomBridge}[2]{%
  \begin{tikzpicture}[
    every node/.style={inner sep=0pt, outer sep=0pt},
    rowlabel/.style={anchor=east, font=\scriptsize, align=right},
    collabel/.style={anchor=south, font=\scriptsize},
    zlabel/.style={anchor=south, font=\scriptsize\itshape},
    instr/.style={anchor=south, font=\itshape\small, align=center,
                  text width=4.6cm},
    gtframe/.style={draw=black!35, line width=0.5pt, inner sep=0pt},
    zoomline/.style={line width=0.5pt, dashed},
  ]
    \foreach \c/\t in {0/1, 1/7, 2/13, 3/19, 4/25} {%
      \pgfmathsetlengthmacro{\xcol}{\c*(\zqw + \zqsep)}%
      \node[collabel] at (\xcol, 0.5\zqh + 0.04cm) {$t=\t$};
    }

    \pgfmathsetlengthmacro{\yrow}{0pt}%
    \pgfmathsetlengthmacro{\labelx}{-0.5*\zqw - 0.08cm}%
    \foreach \c/\fname in {0/0001, 1/0007, 2/0013, 3/0019, 4/0025}{%
      \pgfmathsetlengthmacro{\xcol}{\c*(\zqw + \zqsep)}%
      \node at (\xcol, \yrow) {%
        \includegraphics[width=\zqw, height=\zqh]{#2/gt/episode00010/color/\fname_color.png}};
    }

    \pgfmathsetlengthmacro{\gc}{-1.5*(\zqh+\zqrowsep)}%
    \pgfmathsetlengthmacro{\xz}{4*(\zqw+\zqsep) + 0.5*\zqw + \zoomgap + 0.5*\zoomw}%
    \foreach \r/\modellabel/\modeldir/\modelfolder/\modelcol in {%
        1/{\Method{}}/{slotdit}/{img_10_psnr=22.68_lpips=0.076}/{green!55!black},
        2/{DiT +\\ SD-VAE}/{sdvae}/{orig_psnr=20.94_lpips=0.112}/{orange!85!black},
        3/{DiT +\\ VideoVAE}/{videovae}/{orig_psnr=20.08_lpips=0.118}/{violet!80!black}%
    }{%
      \pgfmathsetlengthmacro{\yrow}{-\r*(\zqh + \zqrowsep)}%
      \pgfmathsetlengthmacro{\labelx}{(\zqw+\zqsep) - 0.5*\zqw - 0.08cm}%
      \node[rowlabel] at (\labelx, \yrow) {\modellabel};
      \foreach \c/\fname in {1/05, 2/11, 3/17}{%
        \pgfmathsetlengthmacro{\xcol}{\c*(\zqw + \zqsep)}%
        \node at (\xcol, \yrow) {%
          \includegraphics[width=\zqw, height=\zqh]%
            {#2/\modeldir/\modelfolder/predicted_frames/pred_\fname.png}};
      }
      \pgfmathsetlengthmacro{\xlast}{4*(\zqw + \zqsep)}%
      \node[draw=\modelcol, line width=0.9pt, inner sep=0pt] (zsrc\r) at (\xlast, \yrow) {%
        \includegraphics[width=\zqw, height=\zqh]%
          {#2/\modeldir/\modelfolder/predicted_frames/pred_23.png}};
      \pgfmathtruncatemacro{\zi}{\r-1}%
      \pgfmathsetlengthmacro{\yz}{\gc - (\zi-1)*(\zoomh+\zoomvsep)}%
      \node[draw=\modelcol, line width=0.9pt, inner sep=0pt] (zdst\r) at (\xz, \yz) {%
        \includegraphics[width=\zoomw, height=\zoomh]%
          {#2/\modeldir/\modelfolder/predicted_frames/pred_23.png}};
      \draw[zoomline, \modelcol] (zsrc\r.north east) -- (zdst\r.north west);
      \draw[zoomline, \modelcol] (zsrc\r.south east) -- (zdst\r.south west);
    }

    \pgfmathsetlengthmacro{\xmid}{2.*(\zqw + \zqsep)}%
    \node[instr] at (\xmid, 0.5\zqh + 0.32cm) {``#1''};
  \end{tikzpicture}%
}

\begin{figure*}[t]
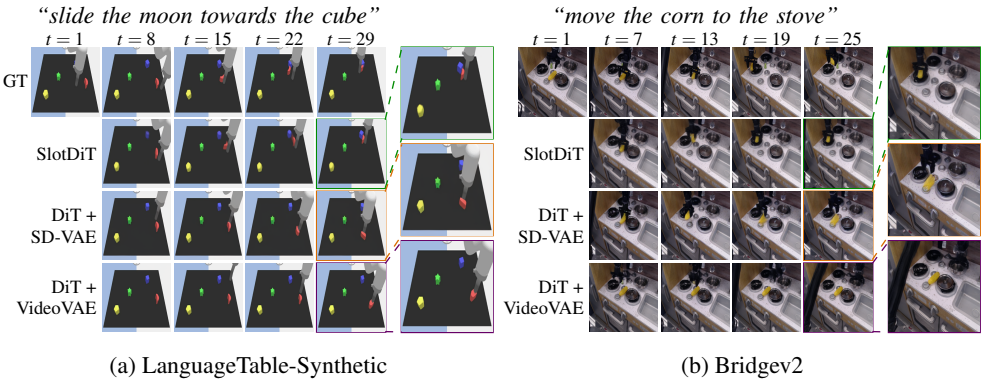

  \centering
  \begin{tabular}{@{}c@{\hspace{0.13cm}}c@{}}
    \videoGenZoomLTSyn%
      {slide the moon towards the cube}%
      {images/videogen_qualitative_syn/ltsyn}%
      {8013}
    &
    \videoGenZoomBridge%
      {move the corn to the stove}%
      {images/videogen_qualitative_real/bridge}
    \\[0.10cm]
    \small (a) \ltsyn{} & \small (b) \bridge{}
  \end{tabular}
  \caption{%
    Qualitative video generation examples on a synthetic (\ltsyn{}) and a real-world (\bridge{}) dataset. Each block shows the ground-truth sequence (top row)
    and the predictions of \Method{}, \sdvaeDiT{}, and \videovaeDiT{}. 
  }
  \vspace{-0pt}
  \label{fig:videogen_qualitative_zoom}
\end{figure*}

\vspace{-0pt}
\subsubsection{Robot Control}
\vspace{-0pt}
Table~\ref{tab:robot_control} evaluates \Method{} and the baselines on closed-loop \ltsyn{} and open-loop \cliport{} robot control, using (trained) inverse dynamics models (IDMs) to map predicted latents to actions (Appendix~\ref{appendix:invdyn}).
We compare \Method{} with the object-centric, slot-based Transformer predictor TextOCVP and VAE-based DiT variants.
On Language\allowbreak Table-Synthetic, we also compare against a slot-based Transformer world model \ocwm{}, using results from the original paper for matching settings rather than rerunning the model.
Additionally, we report \gtslots{}, an oracle that plans using slots extracted from ground-truth 
future observations, providing an upper bound on \Method{}'s performance.
Details on the robot control task and the evaluation protocols are provided 
in Appendices~\ref{appendix:robot_control_task} and \ref{appendix:robot_tasks}.

On \ltsyn{}, we evaluate the training configuration---\blockfour{} scenes with the \taskbb{} (\textit{b2b}) instruction template---and several out-of-distribution settings.
The latter include more complex scenes with eight blocks (\blockeight{}) and unseen instructions requiring reasoning about relative object locations (\textit{b2bR} and \textit{b2R}).

In the in-distribution setting, shown shaded in Table~\ref{tab:robot_control}, \Method{} achieves the highest success rate among all learned predictors, substantially outperforming both slot- and VAE-based models and nearly matching the \gtslots{} oracle.
This small gap with the oracle suggests that the generated slot trajectories preserve most of the task-relevant information required for planning, despite being produced entirely from autoregressive model predictions.
Across the out-of-distribution \blockeight{} scenes and unseen instruction templates, \Method{} also consistently achieves the highest success rates, indicating that object-centric latent spaces capture task-relevant scene dynamics beyond the training distribution.

On \cliport{}, \Method{} also achieves the highest success rate among the
evaluated learned predictors.
This result shows that the benefit of object-centric
diffusion predictions also extends to open-loop action execution in another manipulation environment.

Overall, the robot-control experiments reinforce the findings from the
video generation task: object-centric latent spaces produce predictions
that are substantially more useful for downstream manipulation and decision-making tasks than
alternative latent representations.

\begin{table*}[t]
  \centering
  \footnotesize
  \setlength{\tabcolsep}{2pt}
  \renewcommand{\arraystretch}{1.05}
  \begin{minipage}[t]{0.27\textwidth}
    \vspace{0pt}
    \centering
    \textbf{\cliport{}.}\par\vspace{3pt}
    \begin{tabular}{@{}l c@{}}
      \toprule
      & \multicolumn{1}{c}{\emph{PutInBowl}} \\
      \cmidrule(lr){2-2}
      Model & Success \\
      \midrule
      \gtslots{}        & 87.0\% \\
      \midrule
      \ocwm{}           & \TBD \\
      TextOCVP           & \secondbest{50.0\%} \\
      \cmidrule(l){1-2}
      \sdvaeDiT{}       & 5.0\% \\
      \imagevaeDiT{}    & 0.5\% \\
      \vavaeDiT{}       & 9.5\% \\
      \videovaeDiT{}    & 10.0\% \\
      \cmidrule(l){1-2}
      \Method{} (Ours)  & \best{73.0\%} \\
      \bottomrule
    \end{tabular}
  \end{minipage}
  \hfill
  \begin{minipage}[t]{0.70\textwidth}
    \vspace{0pt}
    \centering
    \textbf{\ltsyn{}.}\par\vspace{2pt}
    \begin{tabular}{@{}l >{\columncolor{gray!12}}c c c c c c@{}}
      \toprule
      & \multicolumn{3}{c}{\blockfour{}}
      & \multicolumn{3}{c}{\blockeight{}} \\
      \cmidrule(lr){2-4}\cmidrule(lr){5-7}
      Model & \emph{b2b} & \emph{b2bR} & \emph{b2R}
            & \emph{b2b} & \emph{b2bR} & \emph{b2R} \\
      \midrule
      \gtslots{}    & 75.00\% & 72.50\% & 84.00\% & 55.00\% & 53.00\% & 69.00\% \\
      \midrule
      \ocwm{}       & \secondbest{50.00\%} & 26.5\% & \TBD & \TBD & \TBD & \TBD \\
      TextOCVP      & 43.50\% & \secondbest{28.5\%} & 27.5\% & \secondbest{18.00\%} & 7.00\% & 20.50\% \\
      \cmidrule(l){1-7}
      \sdvaeDiT{}   & 16.00\% & 10.50\% & 17.00\% & 11.50\% & 6.50\% & 16.50\% \\
      \imagevaeDiT{} & 10.50\% & 5.50\% & 13.00\% & 7.50\% & 6.00\% & 9.00\% \\
      \vavaeDiT{}   & 16.50\% & 12.50\% & 20.00\% & 8.50\% & 6.50\% & 16.00\% \\
      \videovaeDiT{} & 21.00\% & 18.00\% & \secondbest{40.00\%} & 13.00\% & \secondbest{10.00\%} & \secondbest{27.00\%} \\
      \cmidrule(l){1-7}
      \Method{} (Ours) & \best{74.50\%} & \best{51.50\%} & \best{59.00\%} & \best{24.00\%} & \best{17.00\%} & \best{34.00\%} \\
      \bottomrule
    \end{tabular}
  \end{minipage}
  \caption{
  	Robot-control success rate (\%)~\up{} for open-loop action execution on
  	\cliport{} and closed-loop control on \ltsyn{}.
  	On \ltsyn{}, the shaded column is the training setting (\blockfour{}~/~\emph{b2b}), while the rest are out-of-distribution settings. 
  	Best two results (excluding oracle) are shown in bold and underlined.
  	For \ocwm{}~\cite{jeong2025object}, '--' indicates settings for which no reported result is available.
  }
  \label{tab:robot_control}
\end{table*}

\vspace{-0pt}
\subsection{Model Analysis}\label{sec:robustness}

\subsubsection{Robustness}
Table~\ref{tab:robot_control} additionally evaluates generalization to
unseen instruction templates (\textit{b2bR}, \textit{b2R}) and a more
challenging scene configuration (\blockeight{}).
Despite these distribution shifts, \Method{} consistently achieves the highest success rates among all compared methods, demonstrating robustness to both novel scene configurations and language instructions.
Additional qualitative examples in these out-of-distribution settings are provided in
Appendix~\ref{appendix:robot_qualitative}.

\subsubsection{Controllability}
A key requirement of text-guided video generation models is the ability
to modify their predictions according to the provided language instructions.
Figure~\ref{fig:controllability_qualitative_zoom} shows qualitative examples of \Method{}'s controllability abilities in \ltsyn{} and \bridge{} datasets. 
Given the same initial observation, \Method{} generates distinct future
trajectories for two different instructions, while preserving the overall scene layout and object identities.
In both examples,  \Method{} is able to identify the newly referenced objects and correctly applies the described motion, seamlessly adapting to the new text instruction. 
Additional examples across all datasets are provided in Appendix~\ref{appendix:controllability}.

\ifdefined\ctrlzimgw\else\newlength{\ctrlzimgw}\fi   \setlength{\ctrlzimgw}{0.88cm}
\ifdefined\ctrlzimgh\else\newlength{\ctrlzimgh}\fi   \setlength{\ctrlzimgh}{0.88cm}
\ifdefined\ctrlzimgsep\else\newlength{\ctrlzimgsep}\fi \setlength{\ctrlzimgsep}{0.04cm}
\ifdefined\ctrlzrowsep\else\newlength{\ctrlzrowsep}\fi \setlength{\ctrlzrowsep}{0.50cm}
\ifdefined\ctrlzcapgap\else\newlength{\ctrlzcapgap}\fi \setlength{\ctrlzcapgap}{0.05cm}
\ifdefined\ctrlzoomw\else\newlength{\ctrlzoomw}\fi   \setlength{\ctrlzoomw}{1.18cm}
\ifdefined\ctrlzoomh\else\newlength{\ctrlzoomh}\fi   \setlength{\ctrlzoomh}{1.18cm}
\ifdefined\ctrlzoomgap\else\newlength{\ctrlzoomgap}\fi \setlength{\ctrlzoomgap}{0.16cm}

\providecommand{\ctrlorigcol}{blue!60!black}
\providecommand{\ctrlchgcol}{orange!85!black}

\providecommand{\controllabilityZoomExample}[4]{%
  \begin{tikzpicture}[
    every node/.style={inner sep=0pt, outer sep=0pt},
    rowlabel/.style={anchor=east, font=\scriptsize, align=right},
    collabel/.style={anchor=south, font=\scriptsize},
    instr/.style={anchor=south, font=\itshape\scriptsize, align=center,
                  text width=4.5cm},
    zoomline/.style={line width=0.5pt, dashed},
  ]
    \pgfmathsetlengthmacro{\yGT}{0cm}%
    \pgfmathsetlengthmacro{\yOrig}{\yGT - \ctrlzimgh - \ctrlzrowsep}%
    \pgfmathsetlengthmacro{\yChang}{\yOrig - \ctrlzimgh - \ctrlzrowsep}%
    \pgfmathsetlengthmacro{\labelxGT}{-0.5*\ctrlzimgw - 0.08cm}%
    \pgfmathsetlengthmacro{\labelxPred}{(\ctrlzimgw+\ctrlzimgsep) - 0.5*\ctrlzimgw - 0.08cm}%
    \pgfmathsetlengthmacro{\xmid}{2*(\ctrlzimgw + \ctrlzimgsep)}%

    \node[collabel] at (0, 0.5\ctrlzimgh + 0.04cm) {$t{=}1$};
    \node[rowlabel] at (\labelxGT, \yGT) {GT};
    \node at (0, \yGT) {%
      \includegraphics[width=\ctrlzimgw, height=\ctrlzimgh]%
        {images/controllability_frames/#1/gt/context_frames/context_00.png}%
    };

    \foreach [count=\colidx from 1] \tlabel/\gtidx/\origname/\changedname in {#4} {%
      \pgfmathsetlengthmacro{\xcol}{\colidx*(\ctrlzimgw + \ctrlzimgsep)}%
      \node[collabel] at (\xcol, 0.5\ctrlzimgh + 0.04cm) {$t{=}\tlabel$};
      \node at (\xcol, \yGT) {%
        \includegraphics[width=\ctrlzimgw, height=\ctrlzimgh]%
          {images/controllability_frames/#1/gt/target_frames/target_\gtidx.png}%
      };
      \node at (\xcol, \yOrig) {%
        \includegraphics[width=\ctrlzimgw, height=\ctrlzimgh]%
          {images/controllability_frames/#1/orig_pred/predicted_frames/\origname}%
      };
      \node at (\xcol, \yChang) {%
        \includegraphics[width=\ctrlzimgw, height=\ctrlzimgh]%
          {images/controllability_frames/#1/changed_prompt_pred/predicted_frames/\changedname}%
      };
      \xdef\lastOrigName{\origname}%
      \xdef\lastChangedName{\changedname}%
    }

    \pgfmathsetlengthmacro{\xlast}{4*(\ctrlzimgw + \ctrlzimgsep)}%
    \pgfmathsetlengthmacro{\xz}{\xlast + 0.5*\ctrlzimgw + \ctrlzoomgap + 0.5*\ctrlzoomw}%
    \node[draw=\ctrlorigcol, line width=0.9pt, inner sep=0pt] (zsrcO) at (\xlast, \yOrig) {%
      \includegraphics[width=\ctrlzimgw, height=\ctrlzimgh]%
        {images/controllability_frames/#1/orig_pred/predicted_frames/\lastOrigName}};
    \node[draw=\ctrlchgcol, line width=0.9pt, inner sep=0pt] (zsrcC) at (\xlast, \yChang) {%
      \includegraphics[width=\ctrlzimgw, height=\ctrlzimgh]%
        {images/controllability_frames/#1/changed_prompt_pred/predicted_frames/\lastChangedName}};
    \node[draw=\ctrlorigcol, line width=0.9pt, inner sep=0pt] (zdstO) at (\xz, \yOrig) {%
      \includegraphics[width=\ctrlzoomw, height=\ctrlzoomh]%
        {images/controllability_frames/#1/orig_pred/predicted_frames/\lastOrigName}};
    \node[draw=\ctrlchgcol, line width=0.9pt, inner sep=0pt] (zdstC) at (\xz, \yChang) {%
      \includegraphics[width=\ctrlzoomw, height=\ctrlzoomh]%
        {images/controllability_frames/#1/changed_prompt_pred/predicted_frames/\lastChangedName}};
    \draw[zoomline, \ctrlorigcol] (zsrcO.north east) -- (zdstO.north west);
    \draw[zoomline, \ctrlorigcol] (zsrcO.south east) -- (zdstO.south west);
    \draw[zoomline, \ctrlchgcol] (zsrcC.north east) -- (zdstC.north west);
    \draw[zoomline, \ctrlchgcol] (zsrcC.south east) -- (zdstC.south west);

    \pgfmathsetlengthmacro{\yOrigCap}{\yOrig + 0.5*\ctrlzimgh + \ctrlzcapgap}%
    \node[instr] at (\xmid, \yOrigCap) {``#2''};
    \node[rowlabel, align=right] at (\labelxPred, \yOrig) {Original\\Caption};

    \pgfmathsetlengthmacro{\yChangCap}{\yChang + 0.5*\ctrlzimgh + \ctrlzcapgap}%
    \node[instr] at (\xmid, \yChangCap) {``#3''};
    \node[rowlabel, align=right] at (\labelxPred, \yChang) {Changed\\Caption};
  \end{tikzpicture}%
}

\begin{figure*}[t]
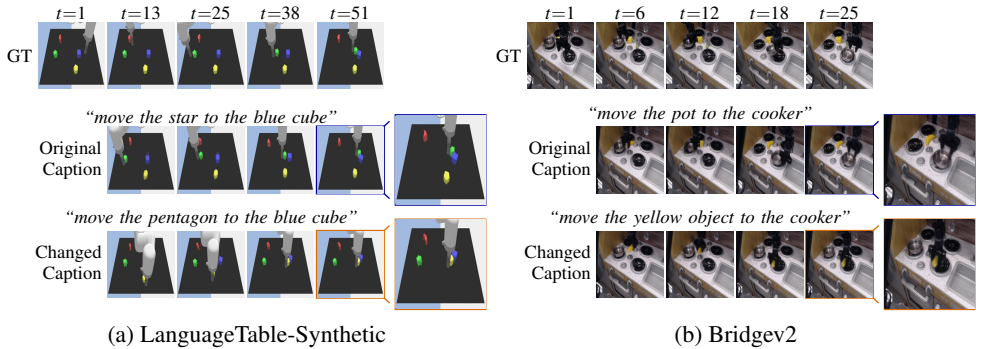

  \centering
  \begin{tabular}{@{}c@{\hspace{0.13cm}}c@{}}
    \controllabilityZoomExample{ltsyn}
      {move the star to the blue cube}
      {move the pentagon to the blue cube}
      {13/12/pred_12.png/pred_12.png,
       25/24/pred_24.png/pred_24.png,
       38/37/pred_37.png/pred_37.png,
       51/50/pred_50.png/pred_50.png}
    &
    \controllabilityZoomExample{bridge}
      {move the pot to the cooker}
      {move the yellow object to the cooker}
      {6/05/pred_05.png/pred_05.png,
       12/11/pred_11.png/pred_11.png,
       18/17/pred_17.png/pred_17.png,
       25/24/pred_24.png/pred_19.png}
    \\[0.0cm]
    \small (a) \ltsyn{} & \small (b) \bridge{}
  \end{tabular}
  \caption{%
    Qualitative controllability examples on \ltsyn{} and \bridge{}. Each block shows the ground-truth video (top row), \Method{}'s rollout conditioned on the
    \emph{original} instruction (middle row), and on a \emph{changed}
    instruction (bottom row). 
  }
  \label{fig:controllability_qualitative_zoom}
  \vspace{-0pt}
\end{figure*}

\begin{table*}[t]
  \centering
  \footnotesize
  \setlength{\tabcolsep}{4pt}
  \renewcommand{\arraystretch}{1.1}
  \begin{tabular}{@{}l c c c c c c c@{}}
    \toprule
    Model
      & Tokens/
      & Encoding
      & Sampling
      & Decoding
      & Total
      & Throughput
      & Speedup \\
      & frame & (s) \dn & (s) \dn & (s) \dn & (s) \dn & (FPS) \up & ($\times$) \up \\
    \midrule
	\sdvaeDiT{}
	& {256}
	& 0.017{\scriptsize$\pm$0.003}
	& 23.01{\scriptsize$\pm$0.29}
	& 0.736{\scriptsize$\pm$0.003}
	& 23.77{\scriptsize$\pm$0.29}
	& 1.64
	& 1.00$\times$ \\
	
	\imagevaeDiT{}
	& {256}
	& \textbf{0.012{\scriptsize$\pm$0.002}}
	& 23.09{\scriptsize$\pm$0.23}
	& 0.232{\scriptsize$\pm$0.001}
	& 23.34{\scriptsize$\pm$0.23}
	& 1.67
	& 1.02$\times$ \\
	
	\vavaeDiT{}
	& {256}
	& 0.019{\scriptsize$\pm$0.003}
	& 22.92{\scriptsize$\pm$0.22}
	& 0.552{\scriptsize$\pm$0.010}
	& 23.49{\scriptsize$\pm$0.22}
	& 1.66
	& 1.01$\times$ \\
	
	\videovaeDiT{}
	& {256}
	& 0.027{\scriptsize$\pm$0.001}
	& \underline{5.21{\scriptsize$\pm$0.09}}
	& 0.562{\scriptsize$\pm$0.002}
	& \underline{5.81{\scriptsize$\pm$0.09}}
	& \underline{6.71}
	& \underline{4.09$\times$} \\
	
	\raeDiT{}
	& {256}
	& \underline{0.015{\scriptsize$\pm$0.002}}
	& 59.23{\scriptsize$\pm$0.54}
	& \textbf{0.135{\scriptsize$\pm$0.001}}
	& 59.38{\scriptsize$\pm$0.54}
	& 0.66
	& 0.40$\times$ \\
	
	\rowcolor{gray!8}	
	\Method{} (Ours)
	& \textbf{10}
	& 0.016{\scriptsize$\pm$0.002}
	& \textbf{3.96{\scriptsize$\pm$0.15}}
	& \underline{0.202{\scriptsize$\pm$0.002}}
	& \textbf{4.18{\scriptsize$\pm$0.15}}
	& \textbf{9.33}
	& \textbf{5.68$\times$} \\
    \bottomrule
  \end{tabular}
  \caption{
  	Inference-time efficiency on \cliport{}, showing
  	wall-clock time (seconds) for generating 39 frames from a single context image.
  	\emph{Speedup} is the runtime improvement relative to \sdvaeDiT{}.
  	Results are averaged over 10 sequences on a single NVIDIA A6000 GPU.
  	\Method{} achieves the highest throughput and lowest overall runtime.
  	Best and second-best results are shown in \best{bold} and \secondbest{underlined}, respectively.
  }
  \label{tab:efficiency}
  \vspace{-0pt}
\end{table*}

\subsubsection{Efficiency}

Table~\ref{tab:efficiency} reports inference-time efficiency on \cliport{}.
\Method{} achieves the highest throughput, generating predictions at
9.33 FPS and providing a 5.68$\times$ speedup over the DiT + SD-VAE
baseline.
The main contributor to this improvement is the diffusion sampling stage, where \Method{} is substantially more efficient than the competing baselines.
This efficiency stems from the compact and semantic object-centric latent space, which reduces the number of tokens processed by the DiT from 256 per frame to only 10 slots.
These results demonstrate that slot-based representations provide a substantially more efficient latent space for diffusion transformers while maintaining strong downstream task performance.

\vspace{-0pt}
\subsubsection{Ablation Studies}
\vspace{0pt}

We conduct ablation studies to validate key design choices in \Method{}.
Table~\ref{tab:ablation_rope} compares the RoPE positional encoding strategies introduced in Section~\ref{sub: diffusion_transformer}: a slot permutation-equivariant, temporal-only variant that assigns the same positional encoding to all slots from the same time step, and a non-equivariant one that is applied along both temporal and slot axes.

\begin{table*}[t]
  \centering
  \footnotesize
  \setlength{\tabcolsep}{1.5pt}
  \renewcommand{\arraystretch}{1.0}
  \begin{tabular}{@{}l ccc ccc ccc ccc@{}}
    \toprule
    RoPE Variant
      & \multicolumn{3}{c}{\cliport{}}
      & \multicolumn{3}{c}{\ltsyn{}}
      & \multicolumn{3}{c}{\ltreal{}}
      & \multicolumn{3}{c}{\bridge{}} \\
    \cmidrule(lr){2-4}\cmidrule(lr){5-7}\cmidrule(lr){8-10}\cmidrule(lr){11-13}
      & LPIPS\dn & JEDi\dn & Succ.\up
      & LPIPS\dn & JEDi\dn & Succ.\up
      & LPIPS\dn & JEDi\dn & Succ.\up
      & LPIPS\dn & JEDi\dn & Succ.\up \\
    \midrule
    Temporal only %
      & \cellcolor{gray!15}\best{0.092} & \cellcolor{gray!15}1.85 & \cellcolor{gray!15}\best{87.8\%}
      & \cellcolor{gray!15}0.131 & \cellcolor{gray!15}\best{1.48} & \cellcolor{gray!15}\best{79.7\%}
      & 0.146 & \best{3.78} & 59.4\%
      & 0.185 & 2.55 & 53.9\% \\
    Temporal + slot %
      & 0.102 & \best{1.63} & 85.6\%
      & \best{0.130} & 1.50 & 75.1\% 
      & \cellcolor{gray!15}\best{0.143} & \cellcolor{gray!15}3.81 & \cellcolor{gray!15}\best{64.6\%}
      & \cellcolor{gray!15}\best{0.180} & \cellcolor{gray!15}\best{2.16} & \cellcolor{gray!15}\best{57.6\%} \\
    \bottomrule
  \end{tabular}
  \caption{
  	Ablation of positional encoding.
  	We compare a slot-permutation-equivariant RoPE (temporal axis only)  against a non-equivariant variant (temporal and slot axes).
  	Metrics are reported at the prediction horizon $\NumPreds=29$, with	\emph{Succ.} 
  	denoting the task-success rate.
  	\colorbox{gray!15}{Highlighted cells} indicate the variant adopted by
  	\Method{} on each dataset.
  }
  \label{tab:ablation_rope}
\end{table*}

\begin{table*}[t]
  \centering
  \footnotesize
  \begin{minipage}[t]{0.309\textwidth}
    \centering
    \footnotesize
    \setlength{\tabcolsep}{1pt}
    \renewcommand{\arraystretch}{1.0}
    \textbf{(a) \bridge{}: \\ \Method{}.}\\[2pt]
    \begin{tabular}{@{}l ccc@{}}
      \toprule
      Variant & LPIPS\dn & JEDi\dn & Succ.\up \\
      \midrule
      \Method{}            & \cellcolor{gray!15}0.180 & \cellcolor{gray!15}2.16 & \cellcolor{gray!15}\best{57.6\%} \\
      \;\,w/ 8 blocks      & 0.179 & 2.55 & 53.5\% \\
      \;\,w/ T5-XL         & 0.181 & 2.22 & 56.9\% \\
      \cmidrule(l){1-4}
      \nonoc			   & 0.347 & 7.30 & 33.0\% \\
      \bottomrule
    \end{tabular}
  \end{minipage}\hfill
  \begin{minipage}[t]{0.331\textwidth}
    \centering
    \footnotesize
    \setlength{\tabcolsep}{0pt}
    \renewcommand{\arraystretch}{1.0}
    \textbf{(b) \ltsyn{}: \Method{}.}\\[2pt]
    \begin{tabular}{@{}l ccc@{}}
      \toprule
      Variant & LPIPS\dn & JEDi\dn & Succ.\up \\
      \midrule
      \Method{}                          & \cellcolor{gray!15}0.131 & \cellcolor{gray!15}1.48 & \cellcolor{gray!15}\textbf{79.7}\% \\
      \;\,w/ DiT-S                       & 0.130 & 1.61 & 68.5\% \\
      \;\,w/ DiT-S + fact.               & 0.163 & 1.45 & 39.6\% \\
      \;\,w/o CA adaLN                   & 0.131 & 1.46 & 72.6\% \\
      \;\,w/ pooling                     & 0.131 & 1.59 & 78.2\% \\
      \cmidrule(l){1-4}
      \nonoc			   & 0.200 & 8.02 & 12.7\% \\
      \bottomrule
    \end{tabular}
  \end{minipage}\hfill
  \begin{minipage}[t]{0.320\textwidth}
    \centering
    \footnotesize
    \setlength{\tabcolsep}{0.5pt}
    \renewcommand{\arraystretch}{1.0}
    \textbf{(c) \ltsyn{}: VAE-based DiT.}\\[2pt]
    \begin{tabular}{@{}l ccc@{}}
      \toprule
      Variant & LPIPS\dn & JEDi\dn & Succ.\up \\
      \midrule
      \sdvaeDiT{}      & \cellcolor{gray!15}0.117 & \cellcolor{gray!15}1.83 & \cellcolor{gray!15}{49.2}\% \\
      \;\,w/ T5-XL     & 0.126 & 2.86 & 36.5\% \\
      \;\,w/ larger dim.   & 0.111 & 1.56 & \textbf{54.3}\% \\
      \;\,w/ DiT-S     & 0.119 & 1.94 & 42.1\% \\
      \cmidrule(l){1-4}
      \vavaeDiT{}      & \cellcolor{gray!15}0.117 & \cellcolor{gray!15}3.80 & \cellcolor{gray!15}\textbf{44.2}\% \\
      \;\,w/ T5-XL     & 0.124 & 3.31 & 34.5\% \\
      \bottomrule
    \end{tabular}
  \end{minipage}
  \caption{%
  	Ablation studies across various datasets and models.
  	Metrics are computed at horizon $\NumPreds=29$, and \emph{Succ.} is the VLM-as-judge task-success rate.}
  \label{tab:ablations_secondary}
  \vspace{-1pt}
\end{table*}

On the simpler synthetic datasets, the permutation-equivariant variant performs better overall and achieves higher task-success rates, 
whereas the	non-equivariant variant performs better on the real-world datasets.
We hypothesize that preserving slot permutation-equivariance provides a	useful inductive bias in simpler environments, while relaxing this constraint offers additional flexibility in more diverse scenes.
Accordingly, we adopt the equivariant variant on synthetic data and the non-equivariant variant on real-world data.

Table~\ref{tab:ablations_secondary} further ablates design choices for \Method{} and the baselines on synthetic and real-world data.
Replacing the T5-small language encoder with the larger T5-XL does not improve performance for either \Method{} or the baselines, likely because the dataset text instructions are relatively simple.
For model capacity, increasing the Diffusion Transformer depth to $\NumPredLayers=14$ on \bridge{} improves over our default $\NumPredLayers=8$, motivating our configuration, while a smaller DiT-S setup ($\NumPredLayers=12$, $\TokenDim=384$, 6 heads) degrades performance on \ltsyn{}.
Increasing the hidden dimensionality $\TokenDim$ improves \sdvaeDiT{}'s performance, highlighting the importance of model capacity for high-dimensional latent spaces.
Nevertheless, its task-success rates remain substantially below those of \Method{}, indicating that the choice of latent representation matters more than modest changes to the DiT architecture.

The non-object-centric (\nonoc{}) ablation replaces the object-centric slots with a single high-dimensional latent vector under the same DiT setup.
\Method{} clearly outperforms this baseline on both datasets, supporting the importance of multi-slot object-centric representations across synthetic and real-world data.

Finally, on \ltsyn{}, replacing spatio-temporal self-attention with factorized attention (fact.), removing adaLN time conditioning from the cross-attention layers, or adding the mean-pooled text embedding to the time embeddings (pooling) all reduce task-success rates, supporting our final architectural choices.

\vspace{-1pt}
\section{Conclusion}

We introduced \Method{}, a text-guided Diffusion Transformer that operates in a compact slot-based latent space for object-centric dynamics and generative modeling. 
Given a reference image and language instruction, \Method{} parses the scene into a set of slots and autoregressively denoises future slot trajectories. 
To study the impact of latent-space design for diffusion transformers, we compared slot-based, VAE-based, and semantics-aligned representations within a unified DiT framework across four simulated and real-world robotic datasets.
\Method{} achieves competitive video-generation quality while consistently attaining higher task-success rates in both text-guided video generation and robot control.
Moreover, the compact slot-based latent space makes \Method{} substantially more efficient than the alternative representations.
Overall, object-centric structure proves to be a powerful inductive bias for diffusion-based generative modeling in robotic environments.

\section*{Acknowledgment}
This work has been partially funded by the Federal
Ministry of Research, Technology and Space of Germany
(BMFTR) under grant no. 01IS22094A WestAI and within
the Robotics Institute Germany, grant no. 16ME0999.
Computational resources were provided by the German AI Service Center WestAI.

\bibliography{referencesGjergj}

\clearpage
\appendix

\section{Limitations and Future Work}

\paragraph{Limitations.}
While \Method{} consistently improves downstream task-success rates across all datasets, it has several limitations.
First, the number of slots is a fixed hyperparameter chosen prior to training. For diverse, multi-environment datasets whose scenes contain a varying number of objects, selecting an appropriate slot count is non-trivial and may waste capacity or under-segment cluttered scenes.
Second, because \Method{} operates in a compact, low-dimensional slot space, the representation prioritizes task-relevant structure over fine appearance detail, and can therefore lag behind high-capacity VAE-based latents in visual quality and temporal consistency.
Finally, our robot-control evaluation is restricted to synthetic datasets such as \ltsyn{} and \cliport{}, as it requires an interactive environment in which the generated actions can be executed; extending this evaluation to additional simulated and real-robot settings would offer a more complete picture of the benefits of object-centric latents.

\paragraph{Future Work.}
Several directions could build on our findings.
A natural next step is to scale \Method{} to richer language instructions and to more diverse, visually complex real-world scenes, where robust object-centric decomposition remains an open challenge.
In addition, our results indicate that object-centric latents provide a strong interface for downstream decision-making, motivating their integration into broader planning, policy-learning, and world-modeling pipelines.
Visual fidelity could be improved further by pairing the slot representations with stronger object-centric decoders.
Finally, mitigating potential error accumulation in long autoregressive rollouts, for instance through improved history guidance or longer context windows, could further strengthen long-horizon prediction and control.

\section{Experimental and Implementation Details}
\label{appendix:hyperparams}

\subsection{Baseline Representation Models}
We detail the latent spaces used by the representation baselines: the pretrained checkpoints we use directly, and the representation models we train ourselves. Unless noted otherwise, each custom-trained model uses the same configuration across all datasets and is trained on $256 \times 256$ frames.

\textit{Pretrained checkpoints.} For \sdvaeDiT{} we use the pretrained \sdvae{} from \texttt{stab\allowbreak ilityai/sd-vae-ft-ema} \cite{rombach2022high}, with a spatial downsampling factor of $8$ and $4$ latent channels. For \vavaeDiT{} we use the pretrained \vavae{} checkpoint \texttt{vavae-image\allowbreak net256-f16d32-dinov2} \cite{yao2025reconstruction}, a DINOv2-aligned VAE with a downsampling factor of $16$ and $32$ latent channels. Both are kept frozen and are used only to encode frames into latents and decode them back.

\textit{Custom \imagevae{}.} For \imagevaeDiT{}, we train from scratch an ImageVAE \cite{kingma2013auto} following the SD-VAE training approach \cite{rombach2022high} with $4$ latent channels and a downsampling factor of 8. We use a learning rate of $5 \times 10^{-5}$.

\textit{\videovae{}.} We implement and train \videovae{} following Open-Sora-Plan \cite{lin2024open} and the official implementation of DFoT \cite{song2025historyguided}. We use a spatial downsampling factor of $8$, a temporal downsampling factor of $4$, $16$ latent channels, and a learning rate of $10^{-4}$. 

\textit{RAE decoder.} For the \raeDiT{} baseline, the DINOv2-ViT-Base-with-registers encoder \cite{darcet2024vision} is frozen, and we train only a ViT-Base decoder to reconstruct images from its features, following RAE \cite{zheng2026diffusion}. The decoder is trained with a reconstruction loss, an LPIPS perceptual loss, and a GAN loss, 
using the Adam optimizer \cite{kingma2014adam} with a learning rate of $10^{-4}$ and a cosine schedule.

\subsection{Robot Control Task}
\label{appendix:robot_control_task}

For the \textit{robot control task}, we apply an inverse dynamics model (\invdyn{}) to the latent trajectories generated by \Method{} to obtain environment actions.

On \ltsyn{}, we adopt an autoregressive, closed-loop rollout. Given a single context slot set and the text embeddings, \Method{} generates the next $\NumBuffer-1$ sets of slots. The \invdyn{} then maps each pair of consecutive slot sets to an action, and the resulting actions are sequentially applied to the environment. Afterwards, the last observation is encoded into slots by the object-centric module and serves as the context for the next generation step. This process is repeated until the task is solved or the maximum number of environment steps is reached; an episode in which the task remains
unsolved is considered a failure.
We use $\NumBuffer=10$ and evaluate 200 episodes per reward type, with a maximum of 200 environment steps per episode.

On \cliport{}, we follow an open-loop approach, adapting to its high-level pick-and-place action space. More specifically, given one context slot set and the text embeddings, \Method{} autoregressively generates $\NumPreds = 70$ future sets of slots. The \invdyn{} takes the context slot set and the final predicted slot set as input and predicts a single pick-and-place action. This action is executed in the environment, and the episode is classified as either a success or a failure. 
We evaluate 200 episodes.

We apply the same evaluation protocols to the VAE-based baselines in both environments, replacing the object-centric slot representation and its associated inverse dynamics model with the corresponding VAE latent representation and representation-specific inverse dynamics model.

\subsection{Robot Control Environments}
\label{appendix:robot_tasks}

On \cliport{}, each scene contains colored blocks and bowls, and the text
instructs the robot arm to place a specific block into a target bowl.
Meanwhile, our robot-control evaluation on \ltsyn{} considers two scene
configurations and three instruction templates, which we describe below.

\textit{Scene configurations.}
\blockfour{} matches the training configuration: each scene contains four
blocks with unique colors and shapes drawn from the training distribution.
\blockeight{} is a held-out configuration containing eight blocks per scene, including
color--shape combinations not seen during training. It therefore tests
robustness to increased visual clutter, novel object combinations, and a
larger number of distractors.

\textit{Instruction templates.} We evaluate three templates: (i) \taskbb{}, the in-distribution template, e.g.\ \emph{``move the [color] block close to the [color] block''}; (ii) \taskbbrl{}, an unseen template that combines a reference block with a relative direction, e.g.\ \emph{``move the [color] block to the [direction] of the [color] block''}; and (iii) \taskbrl{}, an unseen template that grounds only in a relative direction, e.g.\ \emph{``move the [color] block to the [direction] of the board''}. The last two templates are absent from the training set of \ltsyn{} and test generalization of the text-conditioning pathway.

\textit{Success criterion.}
On \ltsyn{}, success is determined by the task-specific reward function.
For \taskbb{}, the Euclidean distance between the two designated blocks must
fall below $0.05$. For \taskbbrl{}, the moved block must lie within the
prescribed directional region relative to the reference block, using a
distance tolerance of $0.04$, while the reference block must not be displaced
by more than $0.05$. For \taskbrl{}, the moved block must fall within $0.10$
of the target board-relative location. 
An episode is counted as successful if its condition is satisfied within the maximum number of environment steps; otherwise, it is recorded as failure.

\textit{\gtslots{} oracle.} \gtslots{} is not a learned future predictor. Instead, it replaces the
predicted future slots with slots obtained by encoding ground-truth future observations generated by the environment's oracle policy.
In both environments, we otherwise retain the evaluation pipeline used for \Method{}.
On \ltsyn{}, at every step of the closed-loop rollout, slots are extracted from the ground-truth future observation by the (frozen) object-centric module, and the \invdyn{} (Sec.~\ref{appendix:invdyn}) maps the resulting consecutive slot sets to actions, which are then applied to the actual environment.
On \cliport{}, the context observation and the final observation produced by an oracle pick-and-place rollout are encoded into slots. The \invdyn{} maps these context and goal slot sets to a single pick-and-place action, which is then executed in the environment.
This isolates the contribution of the learned future-slot predictor from that of the slot extractor and the inverse dynamics model, and thus serves as an upper bound on how well any slot-based predictor can do given our \invdyn{}.

\subsection{Inverse Dynamics Model}
\label{appendix:invdyn}

For the robot-control tasks, we use an inverse dynamics model (\invdyn{})
that predicts the action connecting a pair of latent representations. On \ltsyn{}, the pair
corresponds to consecutive video frames, whereas on \cliport{}, it corresponds
to the initial and final states of a complete pick-and-place trajectory.

We implement \invdyn{} as a Transformer encoder
\cite{Vaswani_AttentionIsAllYouNeed_2017} with four blocks, eight attention
heads, and a token dimension of $256$. A learnable \texttt{[ACT]} token is
prepended to the latent tokens. Through self-attention, this token aggregates
information from both input representations, and its output is decoded by an
MLP into the predicted action. We retain the same overall Transformer architecture across latent
representations and adapt the input tokenization to the representation type.

On \ltsyn{}, we train \invdyn{} using pairs of consecutive frames. The model
predicts the corresponding $2$-dimensional end-effector action and is trained
by minimizing the mean squared error with respect to the ground-truth action.
We optimize the model with Adam for $1000$ epochs using a learning rate of
$1\times10^{-4}$.

On \cliport{}, we train \invdyn{} using the slots extracted from the first and
last frames of each demonstration. The target is a single $6$-dimensional
position-only pick-and-place action. We minimize the mean squared error on the action and 
optimize the model with Adam for $1000$ epochs using a learning rate of $1\times10^{-4}$.

\subsection{Task-Success Metric}

We report a \textit{task-success rate} computed with a \textit{VLM-as-judge}. The vision-language model (VLM) receives a detailed prompt together with a list of frames, including the first, last, and four intermediate generated image frames. The prompt contains the task instruction and guidelines for deciding when the task counts as solved, and the VLM returns a final verdict, SUCCESS or FAILURE, along with its reasoning. 
The task-success rate is the percentage of sequences judged successful. We use Qwen3-VL-30B-A3B-Instruct \cite{bai2025qwen3} as the VLM.

\subsection{Datasets}
\label{appendix:datasets}

\textit{\cliport{}} \cite{shridhar2022cliport} is a synthetic robot manipulation dataset. We focus on the Put-Block-In-Bowl variant, where each scene contains six objects, which can be colored blocks or bowls. The accompanying instruction describes placing a block into a bowl and follows the template ``put the [\textit{color}] block in the [\textit{color}] bowl''. The dataset consists of around 20,000 training and 500 validation sequences.

\textit{Language Table} \cite{lynch2023interactive} is a suite of robotics datasets of tabletop scenes paired with language instructions. We focus on its \textit{\ltsyn{}} and \textit{\ltreal{}} variants. 
\ltsyn{} consists of synthetic simulation data. We train and evaluate the models on the \blockfour{} variant, where the scenes contain four uniquely colored blocks and are paired with a \taskbb{} instruction. It provides 8,000 training and around 200 validation sequences.
\ltreal{} consists of real robot data, where each scene contains eight blocks spanning four colors, with two blocks per color. Its instructions describe several possible tasks: block-to-block, block-to-relative-location, block-to-block-relative-location, and block-to-absolute-location. In total, \ltreal{} contains more than 400,000 episodes, from which we draw a validation subset of 1,000 sequences. As the full training set is large, we reshuffle it at the start of every epoch and train on only its first ${\sim}30{,}000$ sequences, exposing the model to varied data across epochs while keeping each epoch tractable.

\textit{BridgeData V2 (\bridge{})} \cite{walke2023bridgedata} is a large and diverse real robot manipulation dataset, where each episode is annotated with a language caption describing the performed task. We operate on a smaller subset of around 8,300 training and 700 validation sequences, spanning diverse environments such as toy kitchens and tabletops, with various objects and camera viewpoints, and several thousand unique instructions. The tasks are also diverse, ranging from environment manipulation, such as opening or closing drawers and microwaves, to object manipulation, such as moving an object to another object, placing an object into a container, or moving an object to a relative position.
On \bridge{}, we also employ a text caption cleaning and canonicalization procedure that maps similar instructions to a single common one.

For text-guided video generation we evaluate \Method{} on all four datasets, whereas for robot control we focus on \ltsyn{} and \cliport{}. To keep results comparable across prediction horizons, all evaluation sequences contain at least 30 frames.

\subsection{Implementation Details}
\label{appendix:implementation_details}

We train all models using two NVIDIA A6000 (48GB)
GPUs. All models use a pretrained and frozen T5-small text encoder. We operate \Method{} and TextOCVP, which use slot-based representations, on an image resolution of $224 \times 224$, except on \cliport{}, where we follow TextOCVP and use $336 \times 336$. 
At inference, all models are initialized from a single observed context
frame. For video generation and open-loop robot control on \cliport{},
subsequent autoregressive windows use the most recent
$\NumHistoryFrames=4$ predicted latent sets as context. For closed-loop
robot control on \ltsyn{}, each replanning step is initialized from a
single newly observed context frame.

\textit{Object-Centric Module.} We train the object-centric representation learning module on sequences of length $\NumFrames = 5$ for 1000 epochs. We use $\NumSlots = 10$ slots on \cliport{} and LanguageTable\allowbreak-Real, and $\NumSlots = 8$ on \ltsyn{} and \bridge{}. The slot dimension is $\SlotDim = 128$ on the synthetic datasets and $\SlotDim = 256$ on the real-world ones. As the encoder, we use DINOv2-ViT-Base \cite{oquab2023dinov2}, with 12 layers, a patch size of 14, and a feature dimension of $\DimFeats = 768$.

\textit{\nonoc{} ablation.} To isolate the effect of the multi-slot representation, we replace the $\NumSlots=8$ object-centric slots with a single $512$-dimensional latent vector. We train its encoder-decoder module using the same architecture, objectives, and optimization procedure as our object-centric module, changing only the number and dimensionality of the slots. We evaluate this ablation only on \ltsyn{} and \bridge{}. 
Matching the total dimensionality of the multi-slot representation, i.e., using single-vector dimensions of $1024$ on \ltsyn{} and $2048$ on \bridge{}, performed worse. Therefore, we use a latent dimension of $512$ on both datasets.
The subsequent DiT architecture and training setup are identical to those of \Method{}.

\textit{\Method{}.} We train \Method{} on top of the pretrained, frozen object-centric module, using windows of $\NumBuffer = 10$ slot sets. The Transformer backbone has $\NumPredLayers = 8$ blocks, a hidden dimension of $\TokenDim = 512$, and 8-head attention; on \bridge{} we increase this to $\NumPredLayers = 14$ blocks. We use a text-dropout probability of $0.1$, except on \bridge{}, where we use $0.15$ together with a drop-path probability of $0.1$ to counter overfitting. On the synthetic datasets we apply the RoPE variant that preserves slot permutation-equivariance, whereas on the real, more complex datasets we apply the variant that does not. We optimize with Adam \cite{kingma2014adam} and a learning rate of $2 \times 10^{-4}$ on all datasets, except \bridge{}, where we use AdamW \cite{loshchilov2018decoupled} with a learning rate of $1 \times 10^{-4}$ and a weight decay of $0.05$. All models are trained with a cosine-annealing schedule and gradient clipping, for 1500 epochs.

\textit{Diffusion Parameters.} We use a cosine noise schedule with $\NumDiffSteps = 1000$ diffusion steps, the $\Velocity$-prediction parameterization \cite{salimans2022progressive}, and the fused min-SNR loss reweighting \cite{hang2023efficient}. At inference, we sample with $50$ DDIM \cite{song2021denoising} steps and a guidance scale of $1.5$.

\textit{Baselines.} For the baselines, we report only the differences from \Method{}.
TextOCVP uses the same object-centric module as \Method{} and a Transformer predictor with eight blocks (fourteen on \bridge{}), 8-head attention, a context window of ten frames, and a learning rate of $1 \times 10^{-4}$.
The remaining baselines use a DiT identical to \Method{}, with 3D RoPE \cite{su2024roformer} and a learning rate of $1 \times 10^{-4}$ across all datasets. The VAE-based baselines operate on $256 \times 256$ inputs and differ in spatial compression and channel count. \sdvae{} and the custom \imagevae{} yield $32 \times 32 \times 4$ latents (spatial downsampling factor $8$, DiT patch size $2$), \vavae{} yields $16 \times 16 \times 32$ latents (factor $16$, patch size $1$), and the \videovae{} yields $32 \times 32 \times 16$ latents per frame (spatial factor $8$) with an additional temporal downsampling factor of $4$, and using a patch size of $2$. Each of these reduces a frame to $16 \times 16 = 256$ tokens. 
\raeDiT{} encodes inputs with a frozen DINOv2-ViT-Base with registers \cite{darcet2024vision}, yielding $256$ patch tokens, but with a large dimension of $768$. Following RAE \cite{zheng2026diffusion}, we adapt the Transformer backbone to be suitable to train with such embeddings, using a hidden dimension of $\TokenDim = 768$ and $\NumPredLayers = 12$ blocks. We additionally train a ViT-Base decoder for image reconstruction.

\section{Additional Results}

\subsection{Qualitative Video Generation}
\label{appendix:videogen_qualitative}
We provide additional qualitative video-generation examples across all four datasets in Figure~\ref{fig:appendix_videogen_qualitative}, comparing \Method{} against the \sdvaeDiT{} and \videovaeDiT{} baselines. Across both synthetic (\cliport{}, \ltsyn{}) and real-world (\ltreal{}, \bridge{}) settings, \Method{} follows the language instruction and remains faithful to the ground-truth dynamics over long horizons, whereas the VAE-based baselines gradually lose object detail and drift away from the instructed behaviour.

On \cliport{}, \sdvaeDiT{} produces artifacts from the early prediction horizons as the block appears directly on the bowl without even being picked up, as can be seen in its predictions where the gray block remains in place across all predictions. \videovaeDiT{} improves temporal consistency, correctly picks up the gray block, but fails to place it in the correct bowl. On the other hand, \Method{} correctly solves the task. 

On the LanguageTable datasets, \Method{} correctly generates frames that illustrate the motion of the instructed task, while the baselines' generations drift during the longer prediction horizons.

On the complex \bridge{} environment, \Method{} correctly picks up the orange cloth from the washing machine and places it in the basket.

These results illustrate that \Method{}, by leveraging its object-centric latent space, shows superior task-success completion rates compared to the VAE-based variants, further supporting the claims in the main paper.

\newlength{\avqw}\setlength{\avqw}{1.10cm}        %
\newlength{\avqh}\setlength{\avqh}{1.10cm}        %
\newlength{\avqsep}\setlength{\avqsep}{0.04cm}    %
\newlength{\avqrowsep}\setlength{\avqrowsep}{0.05cm} %

\newcommand{\appVGcolorGrid}[6]{%
  \begin{tikzpicture}[
    every node/.style={inner sep=0pt, outer sep=0pt},
    rowlabel/.style={anchor=east, font=\scriptsize, align=right},
    collabel/.style={anchor=south, font=\scriptsize},
    instr/.style={anchor=south, font=\itshape\small, align=center,
                  text width=4.8cm},
  ]
    \foreach \c/\t in {#6}{%
      \pgfmathsetlengthmacro{\xcol}{\c*(\avqw + \avqsep)}%
      \node[collabel] at (\xcol, 0.5\avqh + 0.04cm) {$t=\t$};}
    \pgfmathsetlengthmacro{\labelxgt}{-0.5*\avqw - 0.08cm}%
    \node[rowlabel] at (\labelxgt, 0pt) {GT};
    \foreach \c/\fname in {#4}{%
      \pgfmathsetlengthmacro{\xcol}{\c*(\avqw + \avqsep)}%
      \node at (\xcol, 0pt) {\includegraphics[width=\avqw, height=\avqh]%
        {#2/gt/#3/color/\fname_color.png}};}
    \foreach \r/\modellabel/\modeldir in {%
        1/{\Method{}}/{slotdit},
        2/{DiT +\\ SD-VAE}/{sdvae},
        3/{DiT +\\ VideoVAE}/{videovae}}{%
      \pgfmathsetlengthmacro{\yrow}{-\r*(\avqh + \avqrowsep)}%
      \pgfmathsetlengthmacro{\labelx}{(\avqw+\avqsep) - 0.5*\avqw - 0.08cm}%
      \node[rowlabel] at (\labelx, \yrow) {\modellabel};
      \foreach \c/\fname in {#5}{%
        \pgfmathsetlengthmacro{\xcol}{\c*(\avqw + \avqsep)}%
        \node at (\xcol, \yrow) {\includegraphics[width=\avqw, height=\avqh]%
          {#2/\modeldir/#3/color/\fname_color.png}};}}
    \pgfmathsetlengthmacro{\xmid}{2*(\avqw + \avqsep)}%
    \node[instr] at (\xmid, 0.5\avqh + 0.40cm) {``#1''};
  \end{tikzpicture}%
}

\newcommand{\appVGframeGrid}[6]{%
  \begin{tikzpicture}[
    every node/.style={inner sep=0pt, outer sep=0pt},
    rowlabel/.style={anchor=east, font=\scriptsize, align=right},
    collabel/.style={anchor=south, font=\scriptsize},
    instr/.style={anchor=south, font=\itshape\small, align=center,
                  text width=4.8cm},
  ]
    \foreach \c/\t in {#6}{%
      \pgfmathsetlengthmacro{\xcol}{\c*(\avqw + \avqsep)}%
      \node[collabel] at (\xcol, 0.5\avqh + 0.04cm) {$t=\t$};}
    \pgfmathsetlengthmacro{\labelxgt}{-0.5*\avqw - 0.08cm}%
    \node[rowlabel] at (\labelxgt, 0pt) {GT};
    \foreach \c/\fname in {#4}{%
      \pgfmathsetlengthmacro{\xcol}{\c*(\avqw + \avqsep)}%
      \node at (\xcol, 0pt) {\includegraphics[width=\avqw, height=\avqh]%
        {#2/gt/#3/frame_\fname.png}};}
    \foreach \r/\modellabel/\modeldir in {%
        1/{\Method{}}/{slotdit},
        2/{DiT +\\ SD-VAE}/{sdvae},
        3/{DiT +\\ VideoVAE}/{videovae}}{%
      \pgfmathsetlengthmacro{\yrow}{-\r*(\avqh + \avqrowsep)}%
      \pgfmathsetlengthmacro{\labelx}{(\avqw+\avqsep) - 0.5*\avqw - 0.08cm}%
      \node[rowlabel] at (\labelx, \yrow) {\modellabel};
      \foreach \c/\fname in {#5}{%
        \pgfmathsetlengthmacro{\xcol}{\c*(\avqw + \avqsep)}%
        \node at (\xcol, \yrow) {\includegraphics[width=\avqw, height=\avqh]%
          {#2/\modeldir/#3/frame_\fname.png}};}}
    \pgfmathsetlengthmacro{\xmid}{2*(\avqw + \avqsep)}%
    \node[instr] at (\xmid, 0.5\avqh + 0.40cm) {``#1''};
  \end{tikzpicture}%
}

\newcommand{\appVGltsyn}[2]{%
  \begin{tikzpicture}[
    every node/.style={inner sep=0pt, outer sep=0pt},
    rowlabel/.style={anchor=east, font=\scriptsize, align=right},
    collabel/.style={anchor=south, font=\scriptsize},
    instr/.style={anchor=south, font=\itshape\small, align=center,
                  text width=4.8cm},
  ]
    \foreach \c/\t in {0/1, 1/13, 2/25, 3/38, 4/50}{%
      \pgfmathsetlengthmacro{\xcol}{\c*(\avqw + \avqsep)}%
      \node[collabel] at (\xcol, 0.5\avqh + 0.04cm) {$t=\t$};}
    \pgfmathsetlengthmacro{\labelxgt}{-0.5*\avqw - 0.08cm}%
    \node[rowlabel] at (\labelxgt, 0pt) {GT};
    \foreach \c/\fname in {0/1, 1/13, 2/25, 3/38, 4/50}{%
      \pgfmathsetlengthmacro{\xcol}{\c*(\avqw + \avqsep)}%
      \node at (\xcol, 0pt) {\includegraphics[width=\avqw, height=\avqh]%
        {#2/gt/8008/test_\fname.png}};}
    \pgfmathsetlengthmacro{\yslot}{-1*(\avqh + \avqrowsep)}%
    \pgfmathsetlengthmacro{\labelx}{(\avqw+\avqsep) - 0.5*\avqw - 0.08cm}%
    \node[rowlabel] at (\labelx, \yslot) {\Method{}};
    \foreach \c/\fname in {1/11, 2/23, 3/36, 4/48}{%
      \pgfmathsetlengthmacro{\xcol}{\c*(\avqw + \avqsep)}%
      \node at (\xcol, \yslot) {\includegraphics[width=\avqw, height=\avqh]%
        {#2/slotdit/predicted_frames/pred_\fname.png}};}
    \foreach \r/\modellabel/\modeldir in {%
        2/{DiT +\\ SD-VAE}/{sdvae},
        3/{DiT +\\ VideoVAE}/{videovae}}{%
      \pgfmathsetlengthmacro{\yrow}{-\r*(\avqh + \avqrowsep)}%
      \node[rowlabel] at (\labelx, \yrow) {\modellabel};
      \foreach \c/\fname in {1/13, 2/25, 3/38, 4/50}{%
        \pgfmathsetlengthmacro{\xcol}{\c*(\avqw + \avqsep)}%
        \node at (\xcol, \yrow) {\includegraphics[width=\avqw, height=\avqh]%
          {#2/\modeldir/8008/test_\fname.png}};}}
    \pgfmathsetlengthmacro{\xmid}{2*(\avqw + \avqsep)}%
    \node[instr] at (\xmid, 0.5\avqh + 0.40cm) {``#1''};
  \end{tikzpicture}%
}

\begin{figure*}[t]
  \centering
  \begin{tabular}{@{}c@{\hspace{0.35cm}}c@{}}
    \appVGcolorGrid%
      {put the gray block in the green bowl}%
      {images/appendix_videogen_qualitative/cliport}{episode00015}%
      {0/0001, 1/0018, 2/0036, 3/0054, 4/0072}%
      {1/0017, 2/0035, 3/0053, 4/0071}%
      {0/1, 1/18, 2/36, 3/54, 4/72}
    &
    \appVGltsyn%
      {slide the green star next to the blue cube}%
      {images/appendix_videogen_qualitative/ltsyn}
    \\[0.06cm]
    \small (a) \cliport{} & \small (b) \ltsyn{}
    \\[0.22cm]
    \appVGframeGrid%
      {push the yellow hexagon to the top center of the board}%
      {images/appendix_videogen_qualitative/ltreal}{episode00003}%
      {0/000, 1/008, 2/016, 3/024, 4/031}%
      {1/007, 2/015, 3/023, 4/030}%
      {0/1, 1/9, 2/17, 3/25, 4/32}
    &
    \appVGcolorGrid%
      {put the orange cloth in the basket}%
      {images/appendix_videogen_qualitative/bridge}{episode00006}%
      {0/0001, 1/0009, 2/0017, 3/0025, 4/0033}%
      {1/0008, 2/0016, 3/0024, 4/0032}%
      {0/1, 1/9, 2/17, 3/25, 4/33}
    \\[0.06cm]
    \small (c) \ltreal{} & \small (d) \bridge{}
  \end{tabular}
  \caption{%
    Additional qualitative video-generation results across all four datasets:
    the synthetic \cliport{}~(a) and \ltsyn{}~(b), and the real-world
    \ltreal{}~(c) and \bridge{}~(d). Each block shows the ground-truth sequence
    (top row) and the predictions of \Method{}, \sdvaeDiT{}, and \videovaeDiT{}
    at increasing prediction horizons~$t$; the left-most column ($t{=}1$) is the
    conditioning context frame and is shown for the ground truth only. The
    italic text above each block is the language instruction given to the
    models. \Method{} follows the instruction and stays faithful to the
    ground-truth dynamics over long horizons, whereas the VAE-based baselines
    progressively lose object detail and drift from the intended behaviour.%
  }
  \label{fig:appendix_videogen_qualitative}
\end{figure*}

\subsection{Object-Centric Behaviour}
In Figures~\ref{fig:objectcentric_qualitative} and~\ref{fig:objectcentric_qualitative_seq1}, we illustrate the object-centric behaviour of \Method{} on the complex \bridge{} dataset. \Method{}'s generated frames correctly demonstrate the instructed object interactions. Furthermore, we observe how individual slots represent specific components of the scene, and how \Method{} models the dynamics of the objects through its slot representations.

\newlength{\objimgw}\setlength{\objimgw}{1.85cm}
\newlength{\objimgh}\setlength{\objimgh}{1.85cm}
\newlength{\objimgsep}\setlength{\objimgsep}{0.04cm}
\newlength{\objrowsep}\setlength{\objrowsep}{0.05cm}

\newcommand{\objseqdir}{images/slot_segmentations/seq_3}

\newcommand{\objectCentricExample}{%
  \begin{tikzpicture}[
    every node/.style={inner sep=0pt, outer sep=0pt},
    rowlabel/.style={anchor=east, font=\scriptsize, align=right},
    collabel/.style={anchor=south, font=\scriptsize},
    frame/.style={draw=black!40, line width=0.3pt,
                  minimum width=\objimgw, minimum height=\objimgh,
                  inner sep=0pt},
  ]
    \node[collabel] at (0, 0.5\objimgh + 0.04cm) {Context};
    \foreach \c/\t in {1/5, 2/10, 3/15, 4/20, 5/30} {%
      \pgfmathsetlengthmacro{\xcol}{\c*(\objimgw + \objimgsep)}%
      \node[collabel] at (\xcol, 0.5\objimgh + 0.04cm) {$t=\t$};
    }

    \foreach \r/\rlabel in {%
        0/{GT},
        1/{Predictions},
        2/{Slot\\Masks},
        3/{Object 1},
        4/{Object 2},
        5/{Object 3},
        6/{Object 4}%
    }{%
      \pgfmathsetlengthmacro{\yrow}{-\r*(\objimgh + \objrowsep)}%
      \pgfmathtruncatemacro{\labelcol}{(\r==0) ? 0 : 1}%
      \pgfmathsetlengthmacro{\labelx}{\labelcol*(\objimgw+\objimgsep) - 0.5*\objimgw - 0.08cm}%
      \node[rowlabel] at (\labelx, \yrow) {\rlabel};
    }

    \node[frame] at (0, 0) {\includegraphics[width=\objimgw, height=\objimgh]{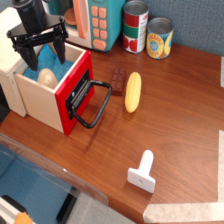}};
    \foreach \c/\f in {1/04, 2/09, 3/14, 4/19, 5/29} {%
      \pgfmathsetlengthmacro{\xcol}{\c*(\objimgw + \objimgsep)}%
      \node[frame] at (\xcol, 0) {\includegraphics[width=\objimgw, height=\objimgh]{\objseqdir/target_frames/target_\f.png}};
    }

    \pgfmathsetlengthmacro{\yrow}{-1*(\objimgh + \objrowsep)}%
    \foreach \c/\f in {1/04, 2/09, 3/14, 4/19, 5/29} {%
      \pgfmathsetlengthmacro{\xcol}{\c*(\objimgw + \objimgsep)}%
      \node[frame] at (\xcol, \yrow) {\includegraphics[width=\objimgw, height=\objimgh]{\objseqdir/predicted_frames/pred_\f.png}};
    }

    \pgfmathsetlengthmacro{\yrow}{-2*(\objimgh + \objrowsep)}%
    \foreach \c/\f in {1/005, 2/010, 3/015, 4/020, 5/030} {%
      \pgfmathsetlengthmacro{\xcol}{\c*(\objimgw + \objimgsep)}%
      \node[frame] at (\xcol, \yrow) {\includegraphics[width=\objimgw, height=\objimgh]{\objseqdir/slot_masks_rgb/frame_\f.png}};
    }

    \foreach \r/\objn in {3/02, 4/00, 5/04, 6/07} {%
      \pgfmathsetlengthmacro{\yrow}{-\r*(\objimgh + \objrowsep)}%
      \foreach \c/\f in {1/005, 2/010, 3/015, 4/020, 5/030} {%
        \pgfmathsetlengthmacro{\xcol}{\c*(\objimgw + \objimgsep)}%
        \node[frame] at (\xcol, \yrow) {\includegraphics[width=\objimgw, height=\objimgh]{\objseqdir/objects/obj_\objn/img_\f.png}};
      }
    }
  \end{tikzpicture}%
}

\begin{figure*}[t]
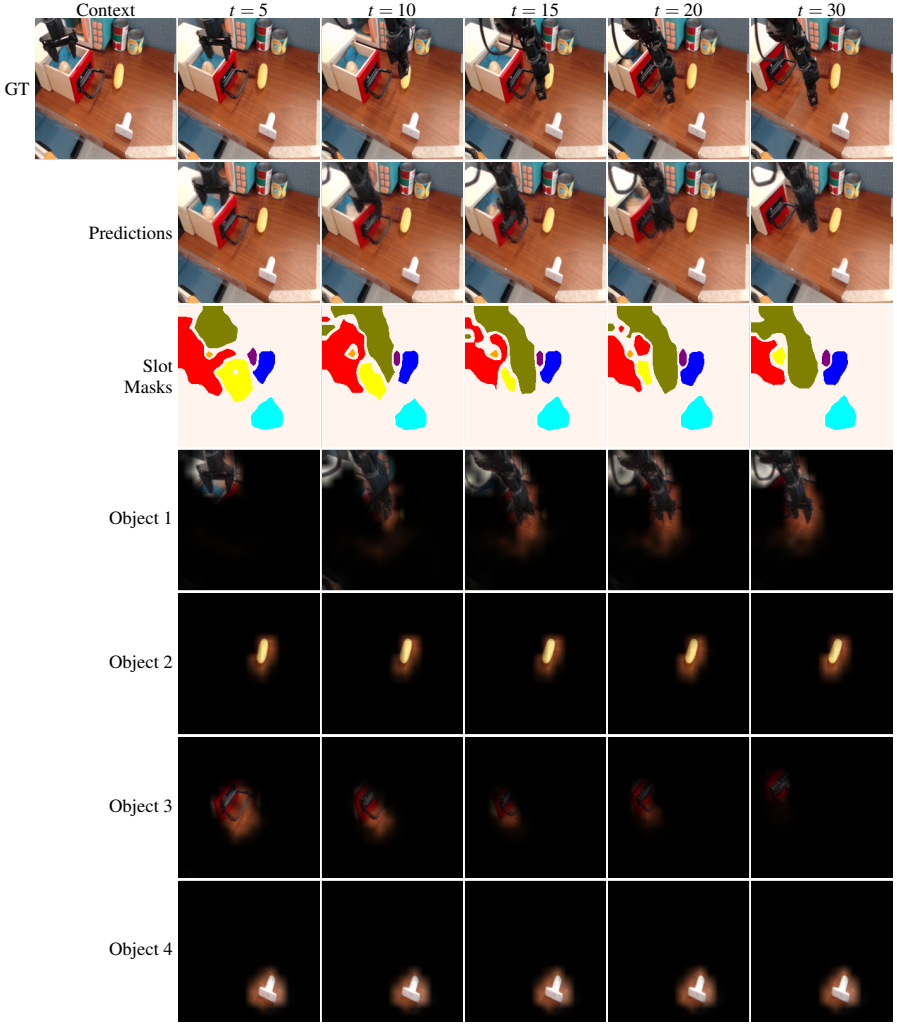

  \centering
  \objectCentricExample
  \caption{%
    Object-centric behaviour of \Method{} on \bridge{}. The first row shows the ground-truth sequence, followed by \Method{}'s predicted
    frames and slot segmentation masks. The rows below display the
    represented objects from four of the predicted slots across various
    time steps, showing how \Method{} models the
    dynamics of individual objects in the scene through slot
    representations.%
  }
  \label{fig:objectcentric_qualitative}
\end{figure*}

\newcommand{\objectCentricExampleSeqOne}{%
  \begin{tikzpicture}[
    every node/.style={inner sep=0pt, outer sep=0pt},
    rowlabel/.style={anchor=east, font=\scriptsize, align=right},
    collabel/.style={anchor=south, font=\scriptsize},
    frame/.style={draw=black!40, line width=0.3pt,
                  minimum width=\objimgw, minimum height=\objimgh,
                  inner sep=0pt},
  ]
    \node[collabel] at (0, 0.5\objimgh + 0.04cm) {Context};
    \foreach \c/\t in {1/5, 2/10, 3/15, 4/20} {%
      \pgfmathsetlengthmacro{\xcol}{\c*(\objimgw + \objimgsep)}%
      \node[collabel] at (\xcol, 0.5\objimgh + 0.04cm) {$t=\t$};
    }

    \foreach \r/\rlabel in {%
        0/{GT},
        1/{Predictions},
        2/{Slot\\Masks},
        3/{Object 1},
        4/{Object 2},
        5/{Object 3},
        6/{Object 4}%
    }{%
      \pgfmathsetlengthmacro{\yrow}{-\r*(\objimgh + \objrowsep)}%
      \pgfmathtruncatemacro{\labelcol}{(\r==0) ? 0 : 1}%
      \pgfmathsetlengthmacro{\labelx}{\labelcol*(\objimgw+\objimgsep) - 0.5*\objimgw - 0.08cm}%
      \node[rowlabel] at (\labelx, \yrow) {\rlabel};
    }

    \node[frame] at (0, 0) {\includegraphics[width=\objimgw, height=\objimgh]{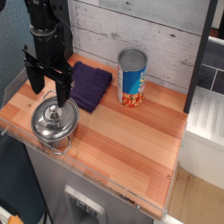}};
    \foreach \c/\f in {1/04, 2/09, 3/14, 4/19} {%
      \pgfmathsetlengthmacro{\xcol}{\c*(\objimgw + \objimgsep)}%
      \node[frame] at (\xcol, 0) {\includegraphics[width=\objimgw, height=\objimgh]{images/slot_segmentations/seq_1/target_frames/target_\f.png}};
    }

    \pgfmathsetlengthmacro{\yrow}{-1*(\objimgh + \objrowsep)}%
    \foreach \c/\f in {1/04, 2/09, 3/14, 4/19} {%
      \pgfmathsetlengthmacro{\xcol}{\c*(\objimgw + \objimgsep)}%
      \node[frame] at (\xcol, \yrow) {\includegraphics[width=\objimgw, height=\objimgh]{images/slot_segmentations/seq_1/predicted_frames/pred_\f.png}};
    }

    \pgfmathsetlengthmacro{\yrow}{-2*(\objimgh + \objrowsep)}%
    \foreach \c/\f in {1/005, 2/010, 3/015, 4/020} {%
      \pgfmathsetlengthmacro{\xcol}{\c*(\objimgw + \objimgsep)}%
      \node[frame] at (\xcol, \yrow) {\includegraphics[width=\objimgw, height=\objimgh]{images/slot_segmentations/seq_1/slot_masks_rgb/frame_\f.png}};
    }

    \foreach \r/\objn in {3/03, 4/04, 5/05, 6/07} {%
      \pgfmathsetlengthmacro{\yrow}{-\r*(\objimgh + \objrowsep)}%
      \foreach \c/\f in {1/005, 2/010, 3/015, 4/020} {%
        \pgfmathsetlengthmacro{\xcol}{\c*(\objimgw + \objimgsep)}%
        \node[frame] at (\xcol, \yrow) {\includegraphics[width=\objimgw, height=\objimgh]{images/slot_segmentations/seq_1/objects/obj_\objn/img_\f.png}};
      }
    }
  \end{tikzpicture}%
}

\begin{figure*}[t]
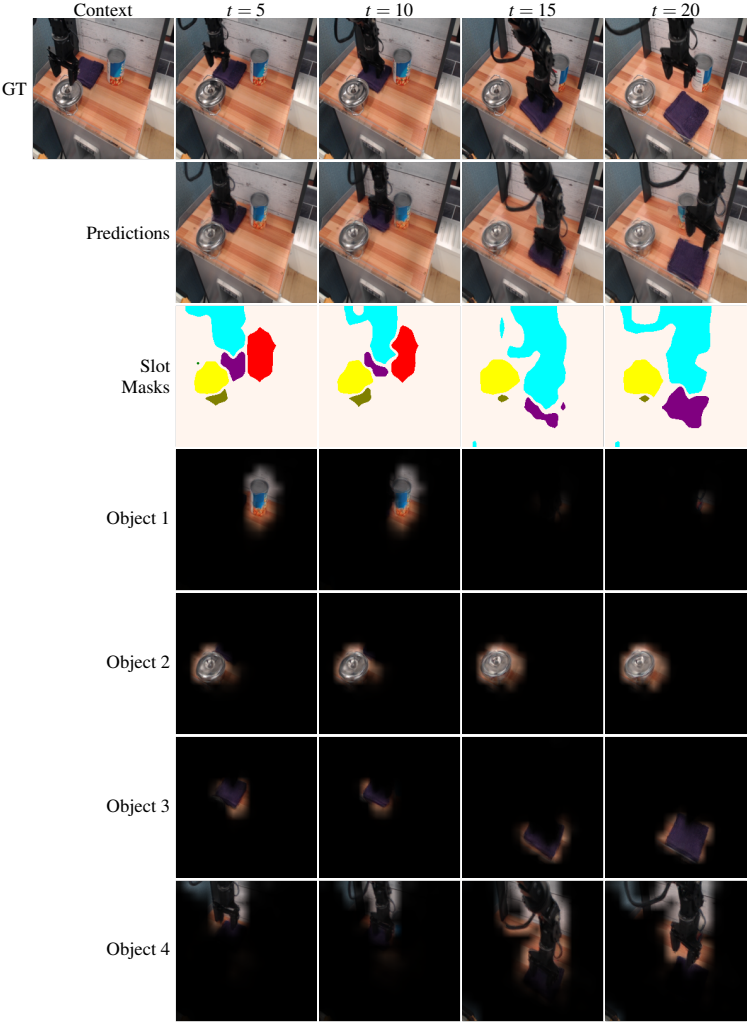

  \centering
  \objectCentricExampleSeqOne
  \caption{%
    Additional visualization of the object-centric behaviour of \Method{} on
    \bridge{}. The rows show the ground-truth sequence, predicted frames, slot
    segmentation masks, and the objects represented by four predicted slots
    across the available prediction horizons.%
  }
  \label{fig:objectcentric_qualitative_seq1}
\end{figure*}

\subsection{Model Robustness}
\label{appendix:robot_qualitative}
In the main paper, we presented quantitative results demonstrating the superior robustness of \Method{} compared to the other baselines in the robot-control setting. Here, we additionally provide qualitative examples from the robot-control task on \ltsyn{}, including scenes with eight blocks and different task templates, as shown in Figure~\ref{fig:robot_control_robustness_qualitative_repr_stacked}.
\Method{} shows superior performance, successfully solving the instructed task across scenes and task templates, while the VAE-based baselines struggle.

\newlength{\rcrobreprLimgw}\setlength{\rcrobreprLimgw}{1.50cm}
\newlength{\rcrobreprLimgh}\setlength{\rcrobreprLimgh}{0.84cm}
\newlength{\rcrobreprLimgsep}\setlength{\rcrobreprLimgsep}{0.05cm}
\newlength{\rcrobreprLrowsep}\setlength{\rcrobreprLrowsep}{0.07cm}

\providecommand{\rcrobsuccess}{\textcolor{green!55!black}{\scriptsize\bfseries S}}
\providecommand{\rcrobfailure}{\textcolor{red!75!black}{\scriptsize\bfseries F}}

\newcommand{\robotControlRobReprExampleLarge}[6]{%
  \begin{tikzpicture}[
    every node/.style={inner sep=0pt, outer sep=0pt},
    rowlabel/.style={anchor=east, font=\small, align=right},
    collabel/.style={anchor=south, font=\small},
    statuslabel/.style={anchor=west},
    instr/.style={anchor=south, font=\itshape\small, align=center,
                  text width=6.6cm},
  ]
    \foreach \t [count=\c from 0] in {#3} {%
      \pgfmathsetlengthmacro{\xcol}{\c*(\rcrobreprLimgw + \rcrobreprLimgsep)}%
      \node[collabel] at (\xcol, 0.5\rcrobreprLimgh + 0.05cm) {$t=\t$};
    }

    \foreach \r/\modellabel/\statustag/\modeldir in {%
        0/{\Method{}}/{#4}/{predictor_dfot_new_fixedRope},
        1/{DiT\\ + \vavae{}}/{#5}/{predictor_dfot_va_vae},
        2/{DiT\\ + \videovae{}}/{#6}/{predictor_dfot_precompVAE_custom_baseline_fixedRope_videoVAE}%
    }{%
      \pgfmathsetlengthmacro{\yrow}{-\r*(\rcrobreprLimgh + \rcrobreprLrowsep)}%
      \pgfmathsetlengthmacro{\labelx}{-0.5*\rcrobreprLimgw - 0.10cm}%
      \node[rowlabel] at (\labelx, \yrow) {\modellabel};
      \foreach \t [count=\c from 0] in {#3}{%
        \pgfmathsetlengthmacro{\xcol}{\c*(\rcrobreprLimgw + \rcrobreprLimgsep)}%
        \node at (\xcol, \yrow) {%
          \includegraphics[width=\rcrobreprLimgw, height=\rcrobreprLimgh]%
            {images/robot_control_frames/#1/\modeldir/c_t\t.png}};
      }
      \pgfmathsetlengthmacro{\statx}{3*(\rcrobreprLimgw + \rcrobreprLimgsep) + 0.5*\rcrobreprLimgw + 0.12cm}%
      \node[statuslabel] at (\statx, \yrow) {\statustag};
    }

    \pgfmathsetlengthmacro{\xmid}{1.5*(\rcrobreprLimgw + \rcrobreprLimgsep)}%
    \node[instr] at (\xmid, 0.5\rcrobreprLimgh + 0.50cm) {``#2''};
  \end{tikzpicture}%
}

\begin{figure*}[p]
  \centering
  \begin{tabular}{@{}c@{}}
    \robotControlRobReprExampleLarge
      {blocktoblock}
      {push the star next to the red block}
      {17,34,51,68}
      {\rcrobsuccess}{\rcrobfailure}{\rcrobfailure}
    \\[0.10cm]
    \small (a) \blockfour{} -- \emph{b2b} (in-distribution)
    \\[0.20cm]
    \robotControlRobReprExampleLarge
      {blocktoblockrelativelocation}
      {move the pentagon to the bottom of the blue block}
      {11,22,33,43}
      {\rcrobsuccess}{\rcrobfailure}{\rcrobfailure}
    \\[0.10cm]
    \small (b) \blockfour{} -- \emph{b2bR} (unseen task)
    \\[0.20cm]
    \robotControlRobReprExampleLarge
      {blocktorelativelocation}
      {slide the moon slightly up and right diagonally}
      {13,25,38,50}
      {\rcrobsuccess}{\rcrobfailure}{\rcrobsuccess}
    \\[0.10cm]
    \small (c) \blockfour{} -- \emph{b2R} (unseen task)
    \\[0.20cm]
    \robotControlRobReprExampleLarge
      {blocktoblock_BLOCK_8}
      {slide the green star next to the green cube}
      {16,32,48,63}
      {\rcrobsuccess}{\rcrobfailure}{\rcrobfailure}
    \\[0.10cm]
    \small (d) \blockeight{} -- \emph{b2b} (unseen scene)
  \end{tabular}
  \caption{%
    Qualitative robot-control robustness of \Method{} on \ltsyn{}, comparing the slot-based predictor against two VAE-based
    DiT baselines: \Method{} (top row of each block), \vavaeDiT{} (middle row),
    and \videovaeDiT{} (bottom row). The four conditions cover the
    in-distribution condition (a), two unseen instruction templates on \blockfour{} scenes (b, c), and the unseen \blockeight{} scene
    configuration (d). The colored tag at the right of each rollout marks the
    closed-loop outcome (\rcrobsuccess{} for success~/~\rcrobfailure{} for failure).%
  }
  \label{fig:robot_control_robustness_qualitative_repr_stacked}
\end{figure*}

\subsection{Controllability}
\label{appendix:controllability}
We provide additional qualitative controllability examples across all four datasets in Figure~\ref{fig:appendix_controllability_qualitative}. For each example, we compare \Method{}'s rollout under the original instruction against its rollout under a changed instruction. \Method{} consistently follows the modified instruction while keeping the rest of the scene consistent with the context frame.

\subsection{Reliability of VLM Judge}
For the text-guided video generation evaluation reported in the main paper, we ask a VLM judge whether each generated sequence completes the instructed task. The task-success rate is the percentage of sequences classified as successful, and we use Qwen3-VL-30B-A3B-Instruct~\cite{bai2025qwen3} as the evaluator.

We further analyse the reliability of this evaluation on the real-world datasets. To this end, we evaluate \Method{} and the main baselines on \ltreal{} and \bridge{} with an additional judge, InternVL3\_5-GPT-OSS-20B-A4B-Preview~\cite{wang2025internvl3}, using the same prompt and following the same protocol.

As shown in Table~\ref{tab:vlm_judge_reliability}, both judges provide broadly consistent conclusions: they produce the same model ranking on \ltreal{} and agree that \Method{} outperforms the VAE-based baselines on both datasets. 
On \ltreal{}, both judges rank \Method{} first and assign it a task-success rate of 64.6\%. 
The only change in ranking occurs on \bridge{}, where the judges reverse the order of \Method{} and TextOCVP: Qwen3-VL assigns 57.6\% to \Method{} and 63.4\% to TextOCVP, whereas InternVL assigns 66.1\% and 62.4\%, respectively. 
Thus, %
both VLM judges tell a largely consistent story and support the overall conclusions of our evaluation. We nevertheless report their scores separately. %

\begin{table*}[t]
	\centering
	\footnotesize
	\setlength{\tabcolsep}{3pt}
	\renewcommand{\arraystretch}{1.0}
	\begin{tabular}{@{}l cc cc@{}}
		\toprule
		Model
		& \multicolumn{2}{c}{\ltreal{}}
		& \multicolumn{2}{c}{\bridge{}} \\
		\cmidrule(lr){2-3}\cmidrule(lr){4-5}
		& InternVL & Qwen3-VL & InternVL & Qwen3-VL \\
		\midrule
		\sdvaeDiT{}    & 56.8\% & 59.0\% & 54.0\% & 49.7\% \\
		\videovaeDiT{} & \secondbest{62.2\%} & \secondbest{61.1\%} & 50.3\% & 40.6\% \\
		TextOCVP        & 48.7\% & 47.4\% & \secondbest{62.4\%} & \best{63.4\%} \\
		\cmidrule(l){1-5}
		\Method{} (Ours) & \best{64.6\%} & \best{64.6\%} & \best{66.1\%} & \secondbest{57.6\%} \\
		\bottomrule
	\end{tabular}
	\caption{Task-success rates (\%)~\up{} assigned by two VLM judges on the real-world datasets. Both judges use the same evaluation protocol. Best results per column are shown in \best{bold} and second-best results in \secondbest{underlined}.}
	\label{tab:vlm_judge_reliability}
\end{table*}

\ifdefined\appctrlimgw\else\newlength{\appctrlimgw}\fi   \setlength{\appctrlimgw}{1.13cm}
\ifdefined\appctrlimgh\else\newlength{\appctrlimgh}\fi   \setlength{\appctrlimgh}{1.13cm}
\ifdefined\appctrlimgsep\else\newlength{\appctrlimgsep}\fi \setlength{\appctrlimgsep}{0.04cm}
\ifdefined\appctrlrowsep\else\newlength{\appctrlrowsep}\fi \setlength{\appctrlrowsep}{0.55cm}
\ifdefined\appctrlcapgap\else\newlength{\appctrlcapgap}\fi \setlength{\appctrlcapgap}{0.05cm}

\providecommand{\appControllabilityExample}[4]{%
  \begin{tikzpicture}[
    every node/.style={inner sep=0pt, outer sep=0pt},
    rowlabel/.style={anchor=east, font=\scriptsize, align=right},
    collabel/.style={anchor=south, font=\scriptsize},
    instr/.style={anchor=south, font=\itshape\scriptsize, align=center,
                  text width=5.5cm},
  ]
    \pgfmathsetlengthmacro{\yGT}{0cm}%
    \pgfmathsetlengthmacro{\yOrig}{\yGT - \appctrlimgh - \appctrlrowsep}%
    \pgfmathsetlengthmacro{\yChang}{\yOrig - \appctrlimgh - \appctrlrowsep}%
    \pgfmathsetlengthmacro{\labelxGT}{-0.5*\appctrlimgw - 0.08cm}%
    \pgfmathsetlengthmacro{\labelxPred}{(\appctrlimgw+\appctrlimgsep) - 0.5*\appctrlimgw - 0.08cm}%
    \pgfmathsetlengthmacro{\xmid}{2*(\appctrlimgw + \appctrlimgsep)}%

    \node[collabel] at (0, 0.5\appctrlimgh + 0.04cm) {$t{=}1$};
    \node[rowlabel] at (\labelxGT, \yGT) {GT};
    \node at (0, \yGT) {%
      \includegraphics[width=\appctrlimgw, height=\appctrlimgh]%
        {images/appendix_controllability_frames/#1/gt/context_frames/context_00.png}%
    };

    \foreach [count=\colidx from 1] \tlabel/\gtidx/\origname/\changedname in {#4} {%
      \pgfmathsetlengthmacro{\xcol}{\colidx*(\appctrlimgw + \appctrlimgsep)}%
      \node[collabel] at (\xcol, 0.5\appctrlimgh + 0.04cm) {$t{=}\tlabel$};
      \node at (\xcol, \yGT) {%
        \includegraphics[width=\appctrlimgw, height=\appctrlimgh]%
          {images/appendix_controllability_frames/#1/gt/target_frames/target_\gtidx.png}%
      };
      \node at (\xcol, \yOrig) {%
        \includegraphics[width=\appctrlimgw, height=\appctrlimgh]%
          {images/appendix_controllability_frames/#1/orig_pred/predicted_frames/\origname}%
      };
      \node at (\xcol, \yChang) {%
        \includegraphics[width=\appctrlimgw, height=\appctrlimgh]%
          {images/appendix_controllability_frames/#1/changed_prompt_pred/predicted_frames/\changedname}%
      };
    }

    \pgfmathsetlengthmacro{\yOrigCap}{\yOrig + 0.5*\appctrlimgh + \appctrlcapgap}%
    \node[instr] at (\xmid, \yOrigCap) {``#2''};
    \node[rowlabel, align=right] at (\labelxPred, \yOrig) {Original\\Caption};

    \pgfmathsetlengthmacro{\yChangCap}{\yChang + 0.5*\appctrlimgh + \appctrlcapgap}%
    \node[instr] at (\xmid, \yChangCap) {``#3''};
    \node[rowlabel, align=right] at (\labelxPred, \yChang) {Changed\\Caption};
  \end{tikzpicture}%
}

\begin{figure*}[t]
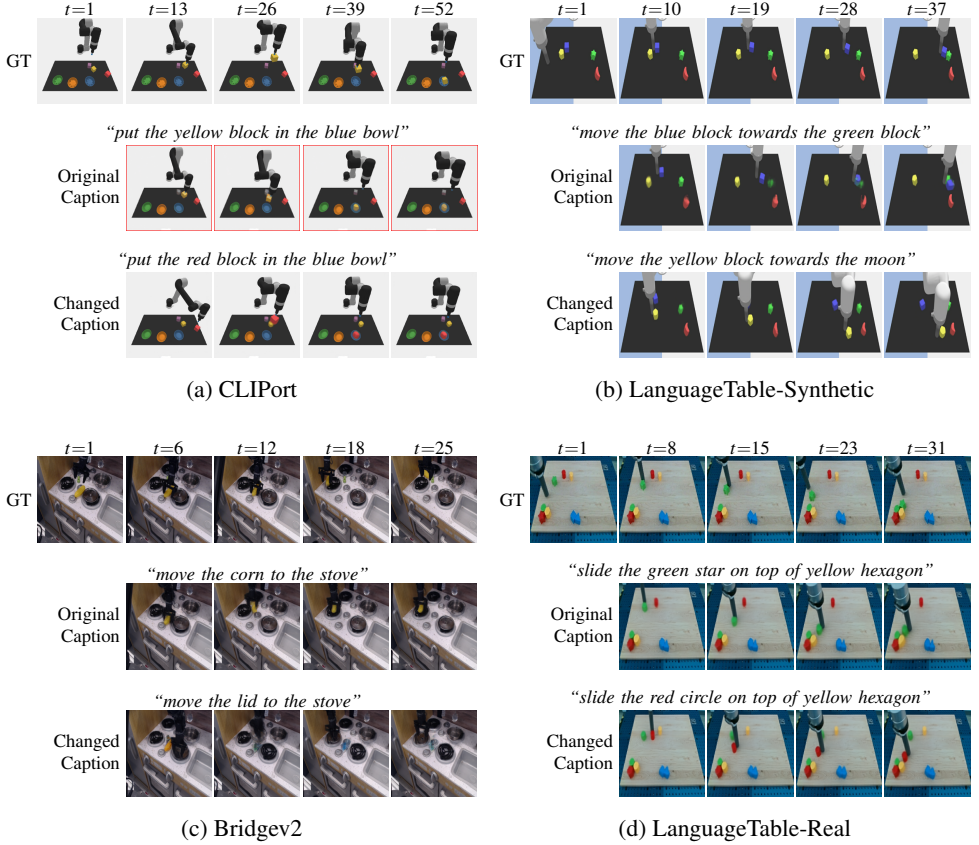

  \centering
  \begin{tabular}{@{}c@{\hspace{0.3cm}}c@{}}
    \appControllabilityExample{cliport}
      {put the yellow block in the blue bowl}
      {put the red block in the blue bowl}
      {13/12/frame_012.png/pred_12.png,
       26/25/frame_025.png/pred_25.png,
       39/38/frame_038.png/pred_38.png,
       52/51/frame_051.png/pred_51.png}
    &
    \appControllabilityExample{ltsyn}
      {move the blue block towards the green block}
      {move the yellow block towards the moon}
      {10/09/pred_09.png/pred_09.png,
       19/18/pred_18.png/pred_18.png,
       28/27/pred_27.png/pred_27.png,
       37/36/pred_36.png/pred_36.png}
    \\[0.10cm]
    \small (a) \cliport{} & \small (b) \ltsyn{}
    \\[0.45cm]
    \appControllabilityExample{bridge}
      {move the corn to the stove}
      {move the lid to the stove}
      {6/05/pred_05.png/pred_05.png,
       12/11/pred_11.png/pred_11.png,
       18/17/pred_17.png/pred_17.png,
       25/24/pred_24.png/pred_24.png}
    &
    \appControllabilityExample{ltreal}
      {slide the green star on top of yellow hexagon}
      {slide the red circle on top of yellow hexagon}
      {8/07/pred_07.png/pred_07.png,
       15/14/pred_14.png/pred_14.png,
       23/22/pred_22.png/pred_22.png,
       31/30/pred_30.png/pred_30.png}
    \\[0.10cm]
    \small (c) \bridge{} & \small (d) \ltreal{}
  \end{tabular}
  \caption{%
    Additional controllability examples across the four datasets. For each
    dataset we show the ground-truth video (top row), \Method{}'s rollout
    conditioned on the \emph{original} instruction (middle row), and on a
    \emph{changed} instruction (bottom row); the italicized text above each
    predicted row is the instruction given to \Method{}. \Method{} follows the
    modified instruction while keeping the rest of the scene consistent with the
    context frame.%
  }
  \label{fig:appendix_controllability_qualitative}
\end{figure*}

\end{document}